\documentclass[10pt,twocolumn,letterpaper]{article}

\usepackage{3dv}
\usepackage{times}
\usepackage{epsfig}
\usepackage{graphicx}
\usepackage{amsmath}
\usepackage{amssymb}
\usepackage{comment}
\usepackage{here}

\newcommand{\fnote}[1]{}

\newcommand{\bnote}[1]{{\color{red} #1 \color{black}}}
\newcommand{\mnote}[1]{}
\newcommand{\onote}[1]{\color{blue} #1 \color{black}}
\newcommand{\snote}[1]{}
\newcommand{\kcut}[1]{}
\newcommand{\ocut}[1]{}
\newcommand{\scut}[1]{}
\newcommand{\jptext}[1]{}

\ifx\pdfoutput\undefined
\else
 \usepackage{CJKutf8}
 \renewcommand{\fnote}[1]{}
 \renewcommand{\mnote}[1]{}
 \renewcommand{\bnote}[1]{}
 \renewcommand{\onote}[1]{}
 \renewcommand{\jptext}[1]{}
\fi

\newcommand{\bm}[1]{\mbox{\boldmath{$#1$}}}

\ifx\etal\undefined\newcommand{\etal}{{\it et al. }}\fi
\ifx\ie\undefined\newcommand{\ie}{{\it i.e.}}\fi
\ifx\eg\undefined\newcommand{\eg}{{\it e.g.}}\fi

\newcommand{\figref}[1]{{Fig.~\ref{fig:#1}}}

\newcommand{\secref}[1]{Sec.~\ref{sec:#1}}
\newcommand{\subsecref}[1]{Sec.~\ref{ssec:#1}}

\usepackage[pagebackref=true,breaklinks=true,letterpaper=true,colorlinks,bookmarks=false]{hyperref}

\threedvfinalcopy 

\def\threedvPaperID{151} 

\ifthreedvfinal\fi
\begin{document}

\title{Unified Underwater Structure-from-Motion}
	
\author{Kazuto Ichimaru$^{\dag}$ \,\,\, Yuichi Taguchi$^{\ddag}$ \,\,\, Hiroshi Kawasaki$^{\dag}$\\
$^{\dag}$ Kyushu University\\
$^{\ddag}$ Mitsubishi Electric Research Labs (MERL)}

\maketitle
\thispagestyle{empty}

\begin{abstract}

This paper shows that accurate underwater 3D shape reconstruction is possible using a single camera, observing a target through a refractive interface. We provide unified reconstruction techniques for a variety of scenarios such as single static camera and moving refractive interface, single moving camera and static refractive interface, and single moving camera and moving refractive interface. In our basic setup, we assume that the refractive interface is planar, and simultaneously estimate the unknown transformations of the planar interface and the camera, and the unknown target shape using bundle adjustment. We also extend it to relax the planarity assumption, which enables us to use waves of the refractive interface for the reconstruction task. Experiments with real data show the superiority of our method to existing methods.

\end{abstract}

\section{Introduction}

Underwater 3D shape reconstruction is a major task in a variety of fields including underwater robot navigation, ocean infrastructure construction, and marine science.
Recent research achieved accurate reconstruction with several techniques such as passive stereo, structured light, and structure-from-motion (SfM).
Among them, SfM, which uses a single camera, is advantageous because of its simple setup and no need of synchronization, 
which also has been popular on the ground. 

\begin{figure}[t]
    \centering
    \includegraphics[width=8cm]{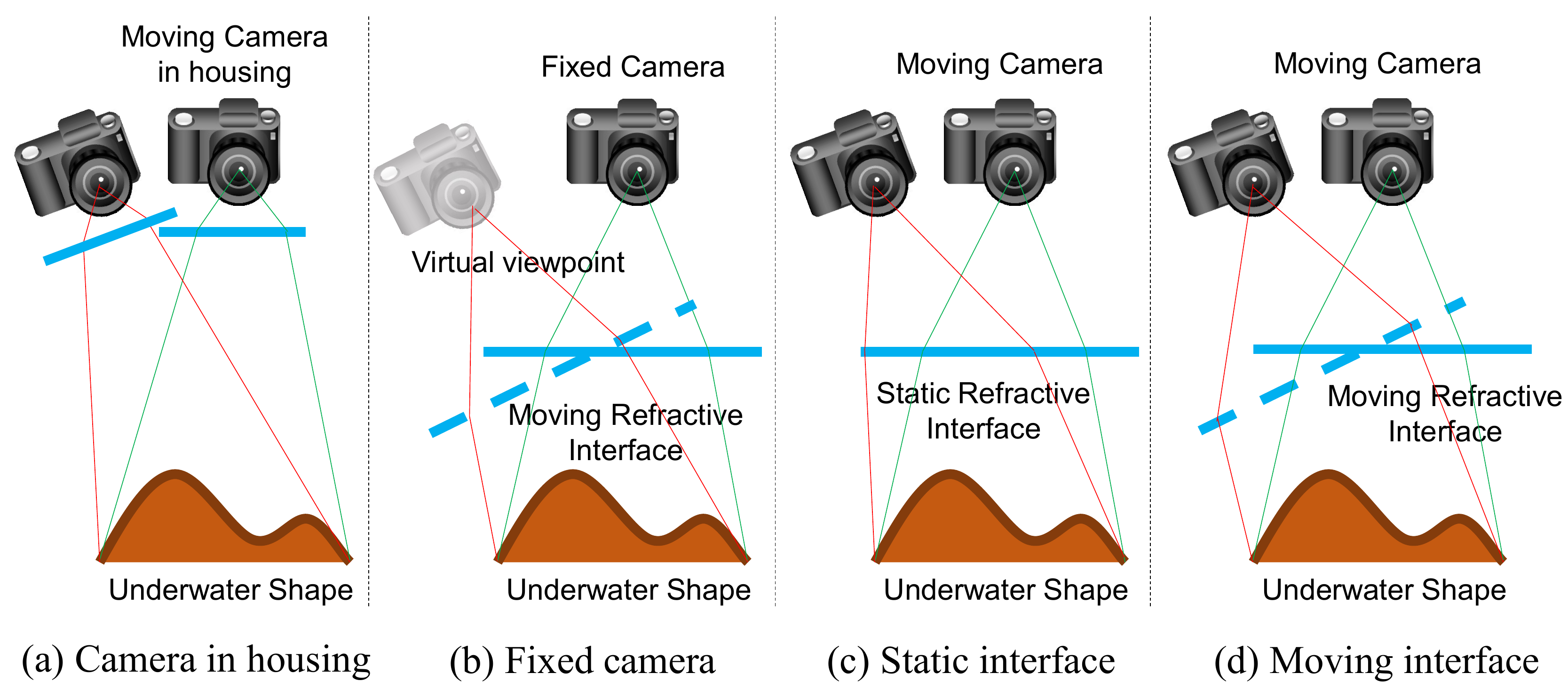}
    \caption{Overview of our geometric settings.}
    \label{fig:geometry_overview}
    \vspace{-0.5cm}
\end{figure}

A typical setup for underwater SfM is shown in \figref{geometry_overview}(a), where a single camera moves together with a planar refractive interface attached to the camera, 
representing a camera in a housing. This setup has been intensively addressed in the literature~\cite{Sedlazeck:ICCV2013,Chadebecq:ICCV2017}.
On the other hand, there are other possible scenarios, which have not been addressed well in the literature, such as single static camera with moving refractive interface 
 (\figref{geometry_overview}(b)), moving camera with static interface (\figref{geometry_overview}(c)), and moving camera with moving interface 
(\figref{geometry_overview}(d)).
The main difference between standard SfM and underwater SfM is the entity of moving object, \ie, only the camera moves in standard SfM, whereas either the camera or the refractive interface, or both of them in the most generalized case shown in \figref{geometry_overview}(d), can move in underwater SfM.

In this paper, we propose a technique to estimate all the unknown parameters for the unified underwater SfM, such as transformations of the camera and the refractive interface as well as underwater scene shapes, 
using an extended bundle adjustment technique. 
In addition, since most of the techniques are under assumption of planarity of the refractive interface, 
we propose a technique to relax the assumption using a soft constraint so that we can apply our technique to 
natural water surface, which can drastically increase the application areas.

The main contributions of this paper are as follows:
\begin{enumerate}
    \item We introduce a unified underwater SfM and propose the solution.
    \item A practical algorithm using extended bundle adjustment for estimating parameters of multiple camera poses, refractive interfaces, and scene structure is provided.
    \item Experimental results using real data show that the proposed method is accurate and stable.
\end{enumerate}

\section{Related work}
Underwater shape reconstruction techniques, such as SfM and multi-view stereo, as well as calibration techniques of camera systems and refractive interface have been widely researched~\cite{Chari:BMVC2009, Agrawal:CVPR2012, Sedlazeck:ECCV2012, Pedersen:CVPRW2018, Ferreira:PRIA2005, Chang:ICCV2013, Sedlazeck:ICCV2013, Shibata:ICRA2015, Chadebecq:ICCV2017, Qian:ECCV2018}.
%
For calibration, Chari and Sturm defined a geometric model for typical underwater camera setup and derived the projection matrix, the fundamental matrix, and the homogprahy matrix~\cite{Chari:BMVC2009}.
Agrawal \etal estimated the transformation of a planar refractive interface using a checkerboard and their analytical forward projection model~\cite{Agrawal:CVPR2012}. 
Sedlazeck and Koch also provided geometric model and optimization process of underwater cameras~\cite{Sedlazeck:ECCV2012}.
Pedersen \etal proposed ray tracing based underwater camera calibration instead of time-consuming forward projection~\cite{Pedersen:CVPRW2018}.

In terms of stereo, Ferreira \etal applied a passive stereo technique to submerged scenes by focal length adjustment and refraction approximation to radial distortion~\cite{Ferreira:PRIA2005}.
Chang and Chen proposed a basic geometric model for underwater multi-view 3D reconstruction with the solution~\cite{Chang:ICCV2013}.
Recently, Qian \etal achieved simultaneous reconstruction of underwater shape and wavy refractive surface using 9 cameras~\cite{Qian:ECCV2018}.
They computed surface normals in two different ways including Snell's law and quadric surface fitting, and applied optimization to make them consistent.

Regarding SfM, 
Sedlazeck and Koch proposed underwater SfM by an efficient optimization equation with a virtual camera model~\cite{Sedlazeck:ICCV2013}.
Shibata \etal achieved scale-reconstructable SfM by using refraction caused by a planar refractive interface~\cite{Shibata:ICRA2015}.
Most recently, Chadebecq \etal considered thick refractive interface for underwater SfM, and derived fundamental matrix~\cite{Chadebecq:ICCV2017}.
All those underwater SfM papers assume the camera in housing scenario.
There are researches which investigate other scenarios on underwater SfM~\cite{Yamashita:IROS2010, Chen:IJCV2013, Alterman:TPAMI2017}.
Yamashita \etal created a special housing that intentionally causes multiple refraction in a single image to apply a passive stereo method, which we can classify as fixed camera scenario~\cite{Yamashita:IROS2010}.
Chen \etal used single camera and transparent glass with known shape for depth estimation and parameter acquisition of the glass using optical flow caused by placing the glass between the camera and the object~\cite{Chen:IJCV2013}.
Alterman \etal introduced a probabilistic model for triangulation in heavy optical distortions caused by refraction, mainly looking up the target in air from underwater~\cite{Alterman:TPAMI2017}.

In terms of shape reconstruction through refractive interface, \eg, water surface, or similar effects, \eg, blur by heat haze, there are several approaches.
Tian \etal reconstructed depth using statistical information of blurred scene by heat haze~\cite{Tian:CVPR2012}, which we use as a comparative method.
They modeled heat haze according to thermodynamics and applied statistical processing to blurred time-series images to obtain depth.
Zhang \etal applied depth-from-defocus (DfD) to underwater scenes, which achieved single-viewpoint simultaneous reconstruction~\cite{Zhang:ECCV2014}.
They derived an analytic formulation of DfD in underwater scenes and normal computation equation.
Their method used a reference image captured with a small aperture, which was a prerequisite of DfD.
Unfortunately, DfD requires such a special optical setting.

There are some work concentrating on refractive interface parameter estimation rather than underwater or background shape reconstruction~\cite{Morris:TPAMI2011, Qian:ECCV2018, Wetzstein:ICCP2011, Kutulakos:IJCV2008, Tanaka:CVPR2017}.
Morris and Kutulakos proposed refraction stereo for reconstructing a wavy liquid surface from known 2D-3D correspondences~\cite{Morris:TPAMI2011}, which was extended by
Qian \etal using a regularizer to improve the accuracy~\cite{Qian:ECCV2018}.
Wetzstein \etal reconstructed static transparent objects and dynamic liquid with light field probes~\cite{Wetzstein:ICCP2011}.
Kutulakos and Steger introduced a triangulation based approach using light field imaging to reconstruct rigid transparent object in air~\cite{Kutulakos:IJCV2008}, and Tanaka \etal extended it to time-of-flight (ToF) based method~\cite{Tanaka:CVPR2017}.
All of the above works assumed a known underwater shape or background to estimate refractive interface parameters, while our goal is to reconstruct an unknown underwater shape.

There are also active methods~\cite{Swirski:ICCV2009, Kawasaki:WACV2012, Anwer:Access2017} for underwater shape reconstruction.
Swirski \etal used caustics of water surface as a pattern projection and applied stereo reconstruction~\cite{Swirski:ICCV2009}.
Kawasaki \etal used a one-shot active technique with a diffractive optical element~\cite{Kawasaki:WACV2012}.
Anwer \etal used a ToF camera and refraction undistortion technique to reconstruct an underwater scene~\cite{Anwer:Access2017}.

Despite many previous work for underwater geometry calibration/reconstruction, most of them assumed specialized geometry settings.
In our work, we consider generalized geometry setting for underwater SfM, which widens the range of applications.

\section{SfM for generalized underwater scenario}
\label{sec:proposal}

\subsection{Geometry definition}
\begin{figure}[t]
    \centering
    \includegraphics[width=8cm]{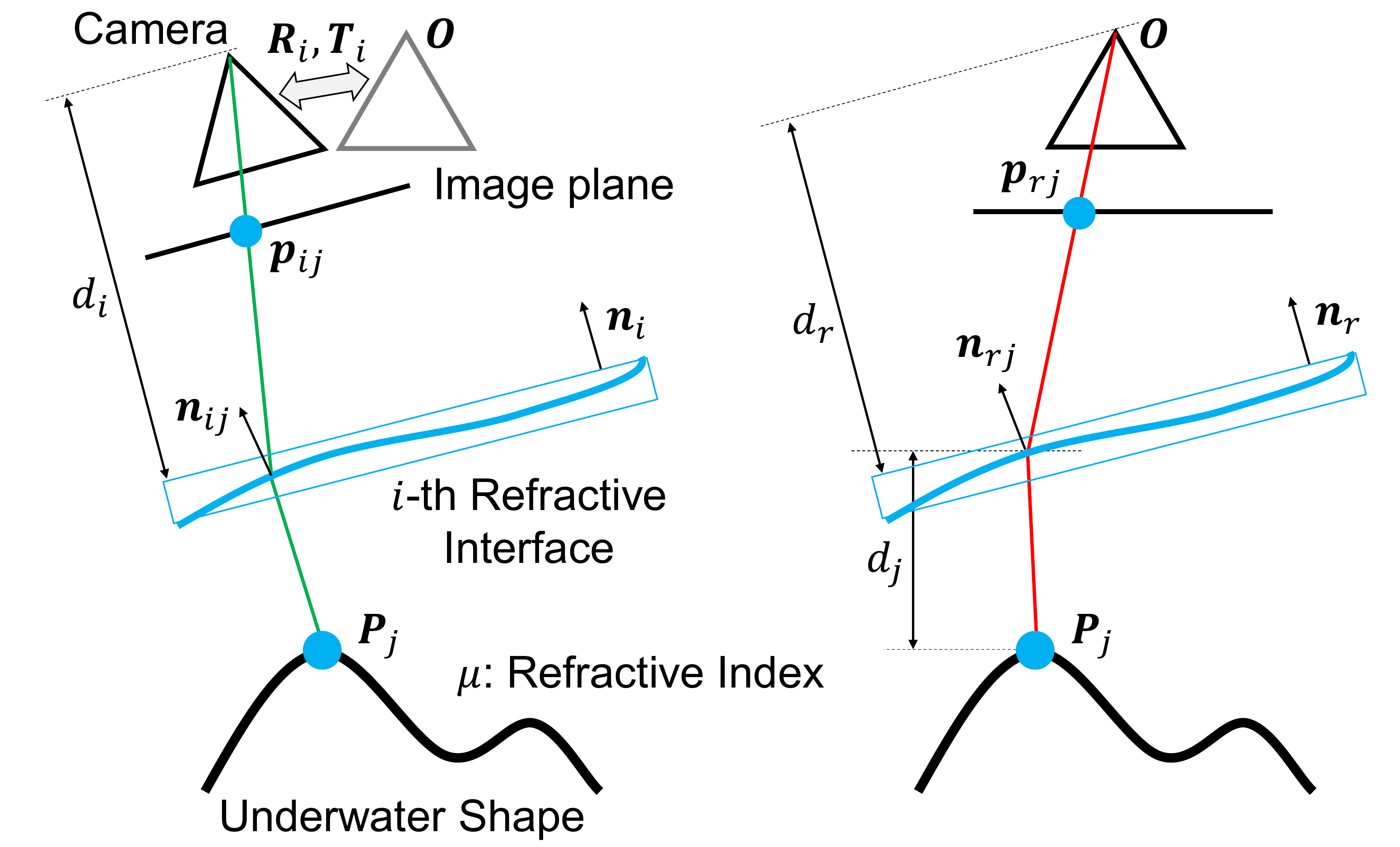}
    \caption{Geometry for capturing underwater images. $\bm{n}_{ij}$ is assigned to each 2D point in each image, which is defined at the intersection between the back-projected ray corresponding to the 2D point and the approximated plane of the refractive interface, represented by $\bm{n}_i$ and $d_i$. Right shows the geometry model for reference image.}
    \label{fig:geometry}
    \vspace{-0.5cm}
\end{figure}

Our setup of generalized underwater scenario consists of a single moving camera, static underwater shape, and moving refractive interface as shown in \figref{geometry}.
The space below the refractive interface is filled with transparent medium with relative refractive index $\mu$.
When we have a set of $I$ images at different viewpoints, transformation of the camera is defined with $\bm{R}_i$ and $\bm{T}_i$ ($i = 1,\ldots,I$).
The refractive interface is assumed to be always between the camera and the underwater shape.
Since we assume the refractive interface is almost planar, we define the transformation of the interface in $i$-th image with depth $d_i$ and normal $\bm{n}_i$ relative to the camera.

When there are $J$ 3D points on the underwater shape and one of the 3D points $\bm{P}_j$ ($j = 1,\ldots,J$) is observed in the $i$-th image at 2D image coordinate $\bm{p}_{ij}$, the ray from the point is refracted on the interface.
If we assume natural interface where the refractive interface is not completely planar, we consider that each point on the refractive interface has its own normal $\bm{n}_{ij}$, which is slightly different from $\bm{n}_i$.
There are no analytical forward projection method for such geometry, since $\bm{n}_{ij}$ depends on the position on the interface.
On the other hand, if the interface is completely planar, we can consider all rays are refracted with normal $\bm{n}_i$, and there is an analytical forward projection method provided in \cite{Agrawal:CVPR2012}.

We assume that $\bm{P}_j$, $\bm{R}_i$, $\bm{T}_i$, $\bm{n}_i$, $d_i$, and $\bm{n}_{ij}$ are unknown, while $\bm{p}_{ij}$, refractive index $\mu$, and camera intrinsic parameters are known.
Our task is to estimate the unknown parameters from the known parameters.

\subsection{Algorithm}
The key observation in developing our algorithm is that, given some estimates of the refractive interface parameters, the back-projection from a 2D point to a 3D point can be easily computed for any interface shapes, while the forward projection from a 3D point to a 2D point can be easily computed only for planar interface shapes.
Thus, we first define a general algorithm that uses back-projection and minimizes the distances between the back-projected rays and 3D points, and then describe a specialized algorithm that uses forward projection and minimizes the reprojection errors when all the refractive interfaces are planar.

First, we choose a single image as a ``reference image'', whose coordinate is treated as the world coordinate. Let $r$-th image be the reference image.
Then, the back-projection of a 2D point $\bm{p}_{rj}$ to a 3D point estimate $\hat{\bm{P}}_{j}$, whose depth is given as $d_j$ relative to the refractive interface, can be described as
\begin{equation}
    \begin{split}
        \hat{\bm{P}}_{j} &= BP(\bm{p}_{rj}, d_j, \bm{n}_r, d_r, \bm{n}_{rj}, \mu) \\
        &= IP(\bm{p}_{rj}, \bm{n}_r, d_r) + RV(\bm{p}_{rj}, \bm{n}_{rj},  \mu) \times d_j,
    \end{split}
    \label{eq:invproj}
\end{equation}
where $IP$ derives the 3D intersection coordinate between the line of sight of $\bm{p}_{rj}$ and the refractive interface $\bm{n}_r, d_r$, and $RV$ derives the vector of the line of sight of $\bm{p}_{rj}$ after refraction.

Next, the back-projection of a 2D point $\bm{p}_{ij}$ for a non-reference image to get the refracted ray vector $\hat{\bm{r}}_{ij}$ and origin $\hat{\bm{o}}_{ij}$ can be described as
\begin{eqnarray}
    \hat{\bm{o}}_{ij} &=& BP(\bm{p}_{ij}, 0, \bm{n}_i, d_i, \bm{n}_{ij}, \mu),\\
    \hat{\bm{r}}_{ij} &=& BP(\bm{p}_{ij}, 1, \bm{n}_i, d_i, \bm{n}_{ij}, \mu) - \hat{\bm{o}}_{ij}.
    \label{eq:invproj_ray}
\end{eqnarray}

Then, we can define a cost function $\bm{D}_{ij}$ to minimize the distance between $\hat{\bm{P}}_{j}$ and the refracted ray as
\begin{eqnarray}
    \hat{\bm{P}}_{ij} &=& (\bm{R}_i \hat{\bm{o}}_{ij} + \bm{T}_i) + t \bm{R}_i \hat{\bm{r}}_{ij},\\
    \bm{D}_{ij} &=& \hat{\bm{P}}_{j} - \hat{\bm{P}}_{ij},
\end{eqnarray}
where $t$ is determined such that $\hat{\bm{P}}_{ij}$ is generated at the same depth as $\hat{\bm{P}}_{j}$.

\begin{figure}[t]
    \centering
    \includegraphics[width=6.8cm]{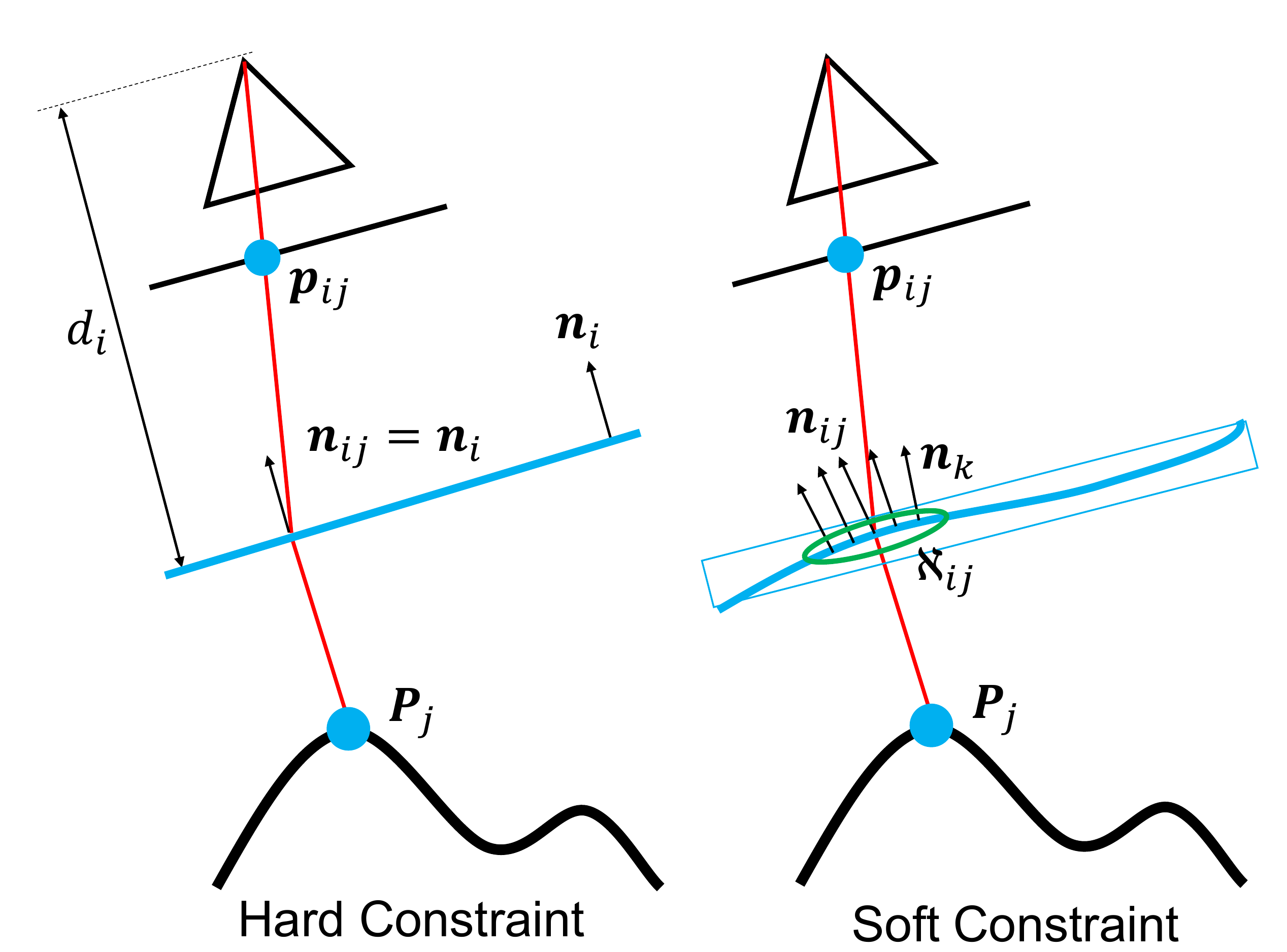}
    \caption{Illustration of the extra constraints. The hard constraint forces all $\bm{n}_{ij}$ to be the same as $\bm{n}_i$, while the soft constraint applies a regularizer to neighborhood normals $\bm{n}_{k}\in\aleph_{ij}$.}
    \label{fig:constraints}
    \vspace{-0.5cm}
\end{figure}

In the general case, the number of unknowns is $9 \times I$ (for $\bm{R}_i$, $\bm{T}_i$, $\bm{n}_i$ and $d_i$) $+$ $J$ (for $d_j$) $+$ $2 \times J \times I$ (for $\bm{n}_{ij}$) $- 6$ (for fixing $\bm{R}_r = \bm{I}$ and $\bm{T}_r = \bm{0}$), while the number of constraints from the cost function is $2 \times J \times (I - 1)$ ($x$ and $y$ coordinate errors for each point for each non-reference image), so this cannot be solved.
Therefore we provide two types of extra constraint: hard constraint and soft constraint (\figref{constraints}).

\paragraph{Hard constraint}
Hard constraint assumes all refractive interfaces are completely planar to fulfill the following equation:
\begin{equation}
    \bm{n}_i = \bm{n}_{ij}.
\end{equation}
Using the hard constraint, the number of unknowns is reduced to $9 \times I + J - 6$, while the number of constraints stays the same as $2 \times J \times (I - 1)$.
It means that the equation can be solved using 12 or more points with 2 images, 7 or more points with 3 images, 6 points with 4 to 8 images, and 5 points with 9 or more images.

When using hard constraint, we can use forward projection of 3D points onto the each images.
Since minimization of reprojection error is preferred to minimization of distance between 3D points in general, we can define another cost function to minimize reprojection error as follows:
\begin{equation}
    \bm{e}_{ij} = FP(\bm{R}_i\hat{\bm{P}}_{j}+\bm{T}_i, \bm{n}_i, d_i, \mu) - \bm{p}_{ij},
\end{equation}
where $FP$ is forward projection function (as for the detail, please refer to \cite{Agrawal:CVPR2012}).

In addition, when the refractive interfaces are completely planar in all images, it is not necessary to specify one image as reference, because all images have the same condition.
Setting a reference image may rather degrade accuracy due to the cost function biased to one specific image.
In order to minimize the reprojection errors equally in all images, a different cost function, which directly optimize 3D point coordinates $\bm{P}_j$, can be defined as follows:
\begin{equation}
   \bm{e}'_{ij} = FP(\bm{R}_i\bm{P}_j+\bm{T}_i, \bm{n}_i, d_i, \mu)-\bm{p}_{ij}.
   \label{eq:cost_noref}
\end{equation}

\paragraph{Soft constraint}
Soft constraint is substantially the same as the hard constraint, \ie, it expects all refractive interfaces to be planar.
However, the soft constraint gives flexibility on each $\bm{n}_{ij}$ by applying a local regularizer, defined as
\begin{equation}
    \sum_{i,j}\sum_{\bm{n}_{k}\in\aleph_{ij}}\|\bm{n}_{k}-\overline{\bm{n}}_{ij}\|^2,
    \label{eq:soft}
\end{equation}
where $\aleph_{ij}$ is a set of neighboring normals to $\bm{n}_{ij}$ in a specific pixel distance, and $\overline{\bm{n}}_{ij}$ is the mean of $\aleph_{ij}$.
This equation means neighboring normals must be the same, but it is better to apply this soft constraint with balancing coefficient when the refractive interface is not completely planar.

\subsection{Geometric constraints for each scenario}
So far we assumed every component of the setup can move (we represented underwater shape as static for simplicity, but it is always relatively static).
We call the most general scenario ``Moving interface''. However, sometimes we can introduce more geometric constraints according to the assumed setup.
We call the scenario ``Static interface'' if the refractive interface is static relative to the underwater shape, and ``Fixed camera'' if the camera is static relative to the underwater shape.
A case with static refractive interface relative to the camera is already studied in \cite{Sedlazeck:ICCV2013, Chadebecq:ICCV2017}, and if all components of the system are static, it is impossible to compute desired parameters due to degeneration.

\paragraph{Static interface}
When the refractive interface is static during capturing, we can consider $\bm{n}_i = \bm{n}_r$ and $d_i = d_r$.
Then, the number of unknowns is $3$ (for $\bm{n}_r$ and $d_r$) $+$ $6 \times (I - 1)$ (for $\bm{R}_i$, $\bm{T}_i$) $+$ $J$ (for $d_j$), while the number of constraints from the reprojection equation remains to be $2 \times J \times (I - 1)$.
It means that the equation can be solved using 15 or more points with 2 images, 7 to 14 points with 3 images, 6 points with 4 images, 5 points with 5 to 7 images, and 4 points with 8 or more images.

\paragraph{Fixed camera}
When the camera is static during capturing, we can consider $\bm{R}_i = \bm{I}$ and $\bm{T}_i = \bm{0}$.
Then, the number of unknowns is $3$ (for $\bm{n}_r$ and $d_r$) $+$ $3 \times I$ (for $\bm{n}_i$ and $d_i$) $+$ $J$ (for $d_j$).
It means that the equation can be solved using 9 or more points with 2 images, 4 to 8 points with 3 images, 3 points with 4 to 8 images, and 2 points with 9 or more images.

Since our moving interface scenario is generalized to all possible situations, it is not necessarily to apply those additional constraints, but they make the solution more accurate and stable, as shown in the experiment.

\subsection{Optimization}
We define the error function $E$ to optimize as follows.
When we use the hard constraint with a reference image,
\begin{equation}
    E_{hard} = \sum_{i,j}\|\bm{e}_{ij}\|^2,
\end{equation}
when we do not specify a reference image,
\begin{equation}
    E_{noref} = \sum_{i,j}\|\bm{e}'_{ij}\|^2,
\end{equation}
and when we use the soft constraint with a reference image,
\begin{equation}
    E_{soft} = \sum_{i,j}\|\bm{D}_{ij}\|^2
      + \lambda\sum_{i,j}\sum_{\bm{n}_{k}\in\aleph_{ij}}\|\bm{n}_{k}-\overline{\bm{n}}_{ij}\|^2,
\end{equation}
where $\lambda$ is a balancing coefficient.
We can choose an error function according to the problem setting.
All unknown parameters are simultaneously optimized to minimize the error function with Levenberg-Marquardt algorithm.

\section{Implementation}
\subsection{Initial value estimation}
In \secref{proposal}, we defined the cost function to optimize.
Appropriate initial values are essential for such an optimization.
We must give initial values to $d_j$, $\bm{R}_i$, $\bm{T}_i$, $\bm{n}_{ij}$, $\bm{n}_i$, $d_i$ when there is a reference image, and
$\bm{P}_j$, $\bm{R}_i$, $\bm{T}_i$, $\bm{n}_i$, $d_i$ when there is no reference image.

The scale ambiguity is a common problem in 3D reconstruction, including our method. To fix the scale, we set $d_r$ to a constant value ($1.0$).
Since the refractive interface such as water surface does not vary its normal and depth significantly, $d_i$ are initialized using the same value as $d_r$.
Also $\bm{n}_r$ and $\bm{n}_i$ are set to $(0, 0, 1)$, since water surface is almost orthogonal to the camera.

\begin{figure}[t]
    \centering
    \includegraphics[width=8.0cm]{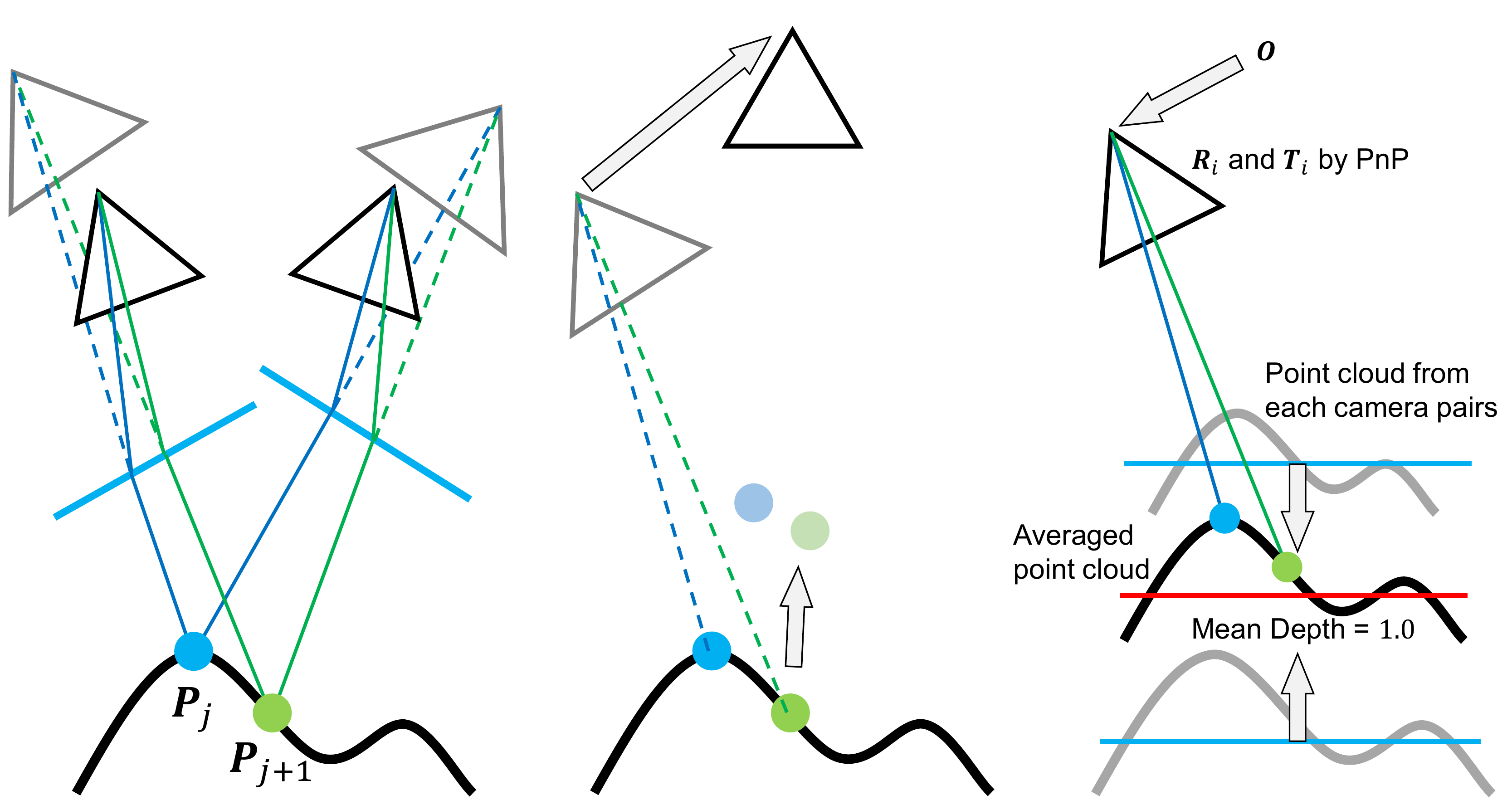}
    \caption{Illustration of the initial 3D coordinate estimation. The virtual viewpoint for each refractive surface is computed by the 5-point algorithm, and then initial 3D coordinates are calculated by triangulation. }
    \label{fig:initial_depth}
    \vspace{-0.5cm}
\end{figure}

Initial value computation for $d_j$, $\bm{P}_j$, $\bm{R}_i$, $\bm{T}_i$, and $\bm{n}_{ij}$ is a more difficult task.
When an approximate depth is known, simply setting $d_j$ to that depth is sufficient, but when even an approximate depth is unknown, we must use a different approach.
Although there are some rough depth estimation methods such as \cite{Tian:CVPR2012} and \cite{Zhang:ECCV2014}, usually they need special settings.
Therefore we use an approach that approximates the geometry by ignoring the refraction, as shown in \figref{initial_depth}(left).
First, two images are selected from captured images.
Since virtual viewpoints are produced because of refraction, we compute the relative transformation between virtual views by estimating the essential matrix from known 2D point correspondences with the 5-point algorithm~\cite{Nister:TPAMI2004} and decomposing it.
Then, we apply triangulation using the estimated transformation to obtain 3D point coordinates.
Despite the fact that the virtual camera appears to have a longer focal length than the actual focal length due to the refraction, we use the originally given camera parameters for triangulation.
The 3D coordinates are reconstructed closer to the camera as shown in \figref{initial_depth} (center), but they are sufficient for initial values.
By repeating the above process for several image pairs, we can get 3D point coordinates for all the observed points.
One remaining problem is scale ambiguity, which means estimated 3D point clouds have different scales since norm of the transformation between the virtual viewpoints is normalized to certain length.
To determine one specific scale, we scale each point cloud so that mean depth of them to be a common constant value $(1.0)$, then average all the point clouds into one point cloud.
After removing outliers for stability, we compute the depth of each points for $d_j$, and relative transformation between virtual viewpoints for $\bm{R}_i$ and $\bm{T}_i$ by solving PnP (\figref{initial_depth}(right)).
The estimated 3D point coordinates are not accurate as we ignore the refraction; however, they are sufficient to be used as initial values.

Once $d_j$ are determined, $\bm{n}_{ij}$ can be calculated immediately according to \eqref{eq:invproj}, or we can just set $\bm{n}_{ij}=\bm{n}_{i}$ when using the hard constraint.

\subsection{Real image processing}
\label{ssec:imageprocessing}
In order to apply our method to real images, it is necessary to calibrate camera intrinsic parameters and extract 2D feature points.
First, we capture several images of a calibration board in air and use Zhang's method~\cite{Zhang:TPAMI2000} to calibrate intrinsic parameters.
Next, we capture a set of underwater images while moving the refractive interface.
Then, image features are extracted from the images, and correspondences are computed across frames (We used SURF~\cite{Bay:ECCV2006} as the feature).
Usually it can be decided which image to use according to the number of detected feature points, or distance of optical flows (longer optical flow makes longer baseline).
Once 2D point coordinates and correspondences between frames are acquired and used frames are decided, our algorithm is applied.

\section{Experiments}
This section describes the experiments we conducted to evaluate our method.
We implemented our algorithm with C++, Python, OpenCV, Eigen, and Ceres solver.

\subsection{Comparative method}
We considered single-viewpoint underwater shape reconstruction techniques as our comparative method.
Refractive DfD~\cite{Zhang:ECCV2014} is such a method, but it requires special optical setting, especially a large aperture.
Since fair quantitative evaluations are difficult with it due to the difference in optical setting, we used depth from optical turbulence (DfT) method~\cite{Tian:CVPR2012} and depth from refraction (DfR) method~\cite{Chen:IJCV2013} instead.
We implemented DfT and DfR in conformity to underwater geometry to use them as comparative methods.

DfT is originally a thermodynamics-based method that exploits heat haze.
If there is a heat source between an object and a camera, heat haze refracts the rays from the object to the camera, and optical turbulence is observed on the captured images.
Since heat haze refracts the rays almost randomly, the turbulence follows Gaussian distribution.
By estimating the parameters of the distribution, the depth of the observed point can be estimated because the farther the point is, the larger the turbulence will be.
In addition, we consider that DfT is intrinsically the same approach as refractive DfD, since only the difference between them is whether the turbulence is observed in the temporal or spatial domain.
Since depth scale and offset information is missing in DfT, we optimized the scale and offset to minimize error between reconstructed shape and ground-truth shape, although it is  impractical for real case.

DfR calculates optical flow between the images with/without refraction, by placing and removing refractive interface.
Then, depth can be analytically computed from the optical flow.
Originally they assumed the scene is basically in air, and flat and thick refractive interface such as glass is placed between the camera and the object.
Since we assume underwater scene, to apply their technique, we put a target object into the water tank, and captured two images with/without water, instead of placing a glass.

In terms of static interface and moving interface scenario, since there is no prior work assuming the same geometry, we compared ours with simple SfM ignoring refraction.
Triangulation in Random Refractive Distortions (TRD)~\cite{Alterman:TPAMI2017} has a similar goal to our work, but their work concentrates on triangulation method given camera poses, while we concentrate on camera pose estimation as well as 3D point reconstruction; thus we consider their approach complementary to our approach.

\subsection{Comparison using flat refractive interface}
\label{ssec:flat}
\begin{figure}[t]
    \centering
    \begin{minipage}{0.49\hsize}
        \includegraphics[width=4cm]{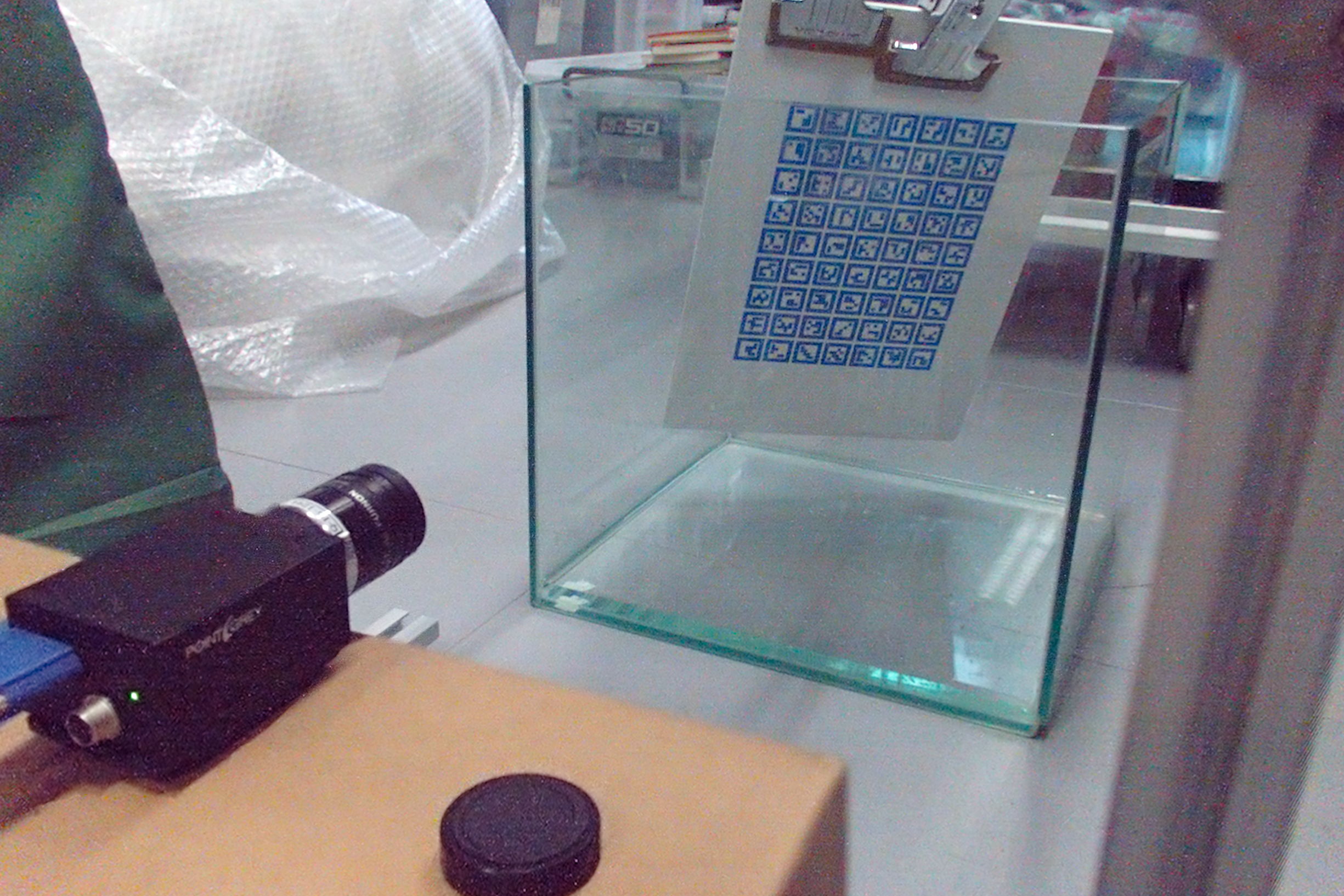}
    \end{minipage}
    \begin{minipage}{0.49\hsize}
        \includegraphics[width=4cm]{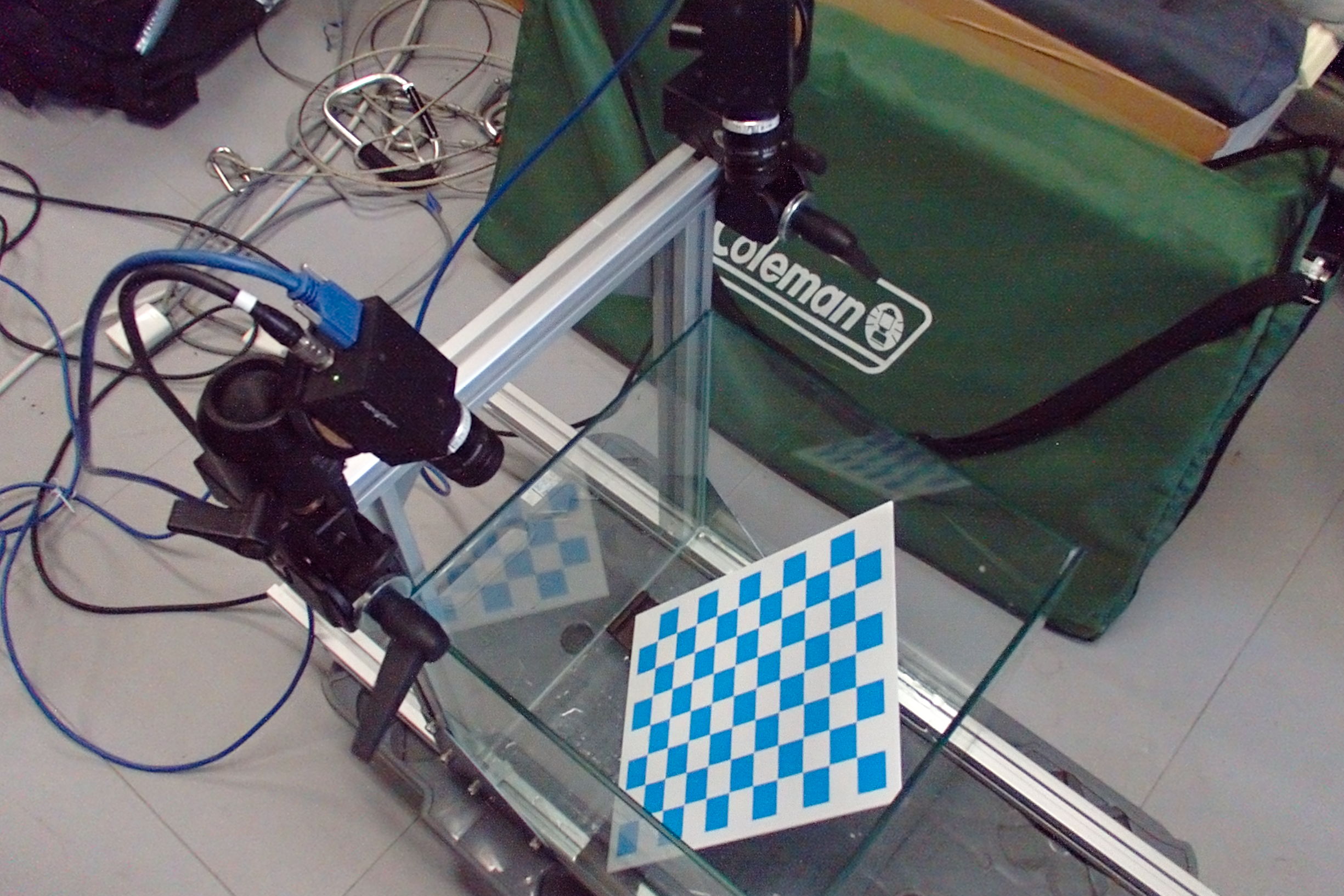}
    \end{minipage}
    \vspace{1mm}
    \caption{{\bf Left: }Appearance of the experimental setup of \subsecref{flat}. {\bf Right: }Appearance of the experimental setup of \subsecref{wave}.}
    \label{fig:expset}
\end{figure}

First, we evaluated the proposed method with completely flat refractive interface, which is an ideal case.
We placed a planar calibration board with markers hanging on an empty water tank, and a camera was placed in front of the water tank.
The water tank could be moved without moving the board and the camera (\figref{expset}(left)).
We first captured images without water to obtain ground-truth (the pose of the calibration board computed from marker information) and as a reference image for DfR.
Then, we poured water into the tank, and captured an image for DfR.
Next, we captured three sets of images: (1) 15 images while moving the water tank for our fixed camera scenario and for DfT; (2) 10 images while moving the camera for our static interface scenario; and (3) 10 images while moving both the camera and the water tank for our moving interface scenario.

\begin{figure}[t]
    \centering
    \includegraphics[width=7cm]{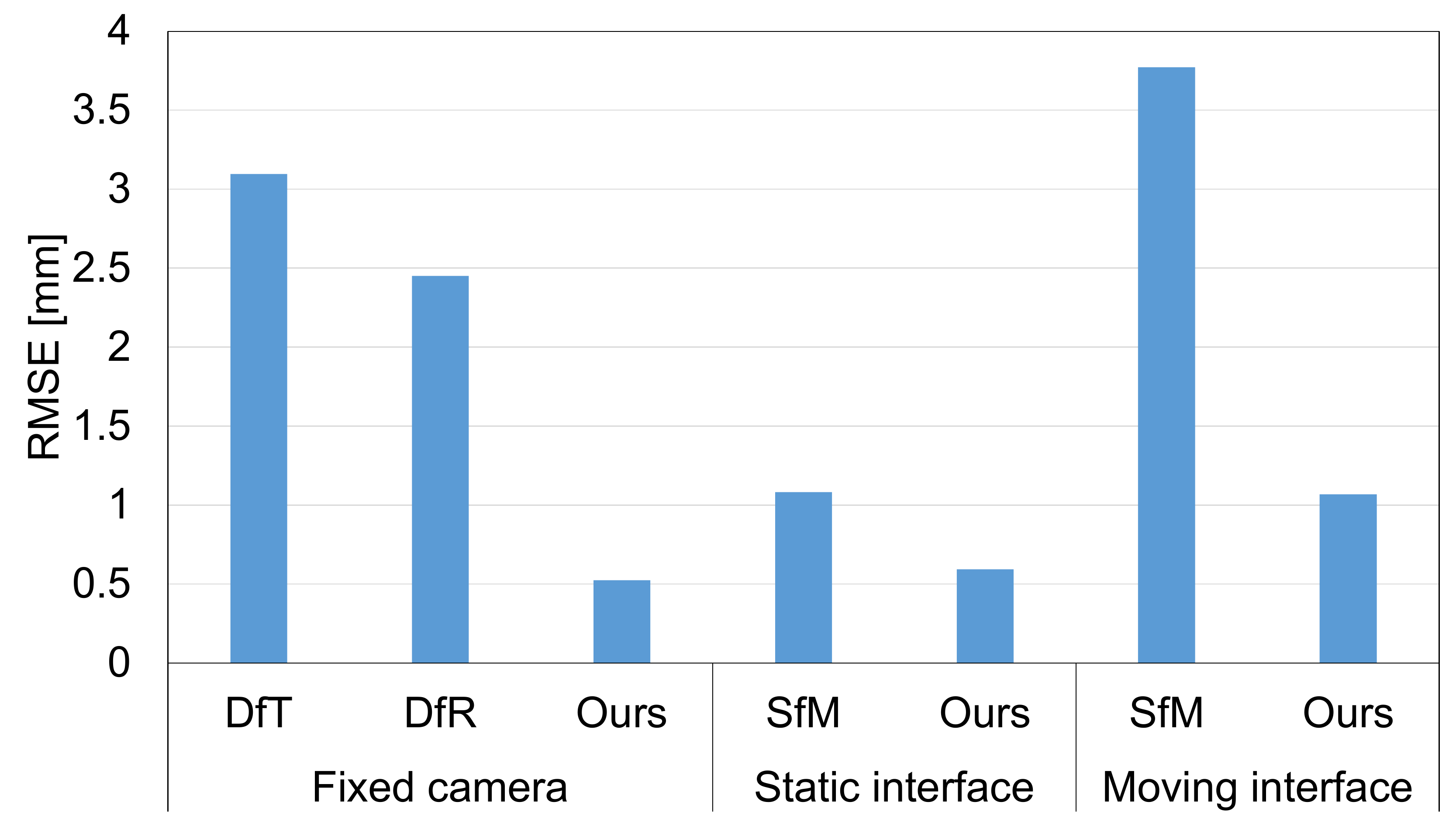}
    \caption{Quantitative result of the experiment with flat refractive interface. Comparative methods are DfT~\cite{Tian:CVPR2012}, DfR~\cite{Chen:IJCV2013}, and simple SfM ignoring refraction.}
    \label{fig:result_flat}
    \vspace{-0.5cm}
\end{figure}

We reconstructed 3D points of the board from the captured images with the proposed method using hard constraint and comparative methods respectively, and computed RMSE from the ground-truth shape after adjusting scales and transformations.
The results are shown in \figref{result_flat}.
For every scenario, the proposed method achieved higher accuracy than the existing methods.
In fixed camera scenario, DfT was stable and successfully estimated approximate 3D point coordinates, however our method performed better.
DfR was also stable for most points despite smaller number of used images, but some points had significantly larger errors due to degeneration, because when comparing two images with/without water, there will be one specific point whose optical flow becomes zero.
In static interface and moving interface scenario, simple SfM produced severe error since it cannot follow the real geometry, while the proposed method achieved millimeter order accuracy.
We noticed our algorithm for moving interface scenario made the error slightly larger than static interface scenario, because the larger number of degrees-of-freedom enlarged the solution space, although it could be useful to represent more complicated geometric setups.

\subsection{Comparison using wavy refractive interface}
\label{ssec:wave}

\begin{figure}[t]
    \centering
    \begin{minipage}{0.49\hsize}
        \centering
        \includegraphics[width=4cm]{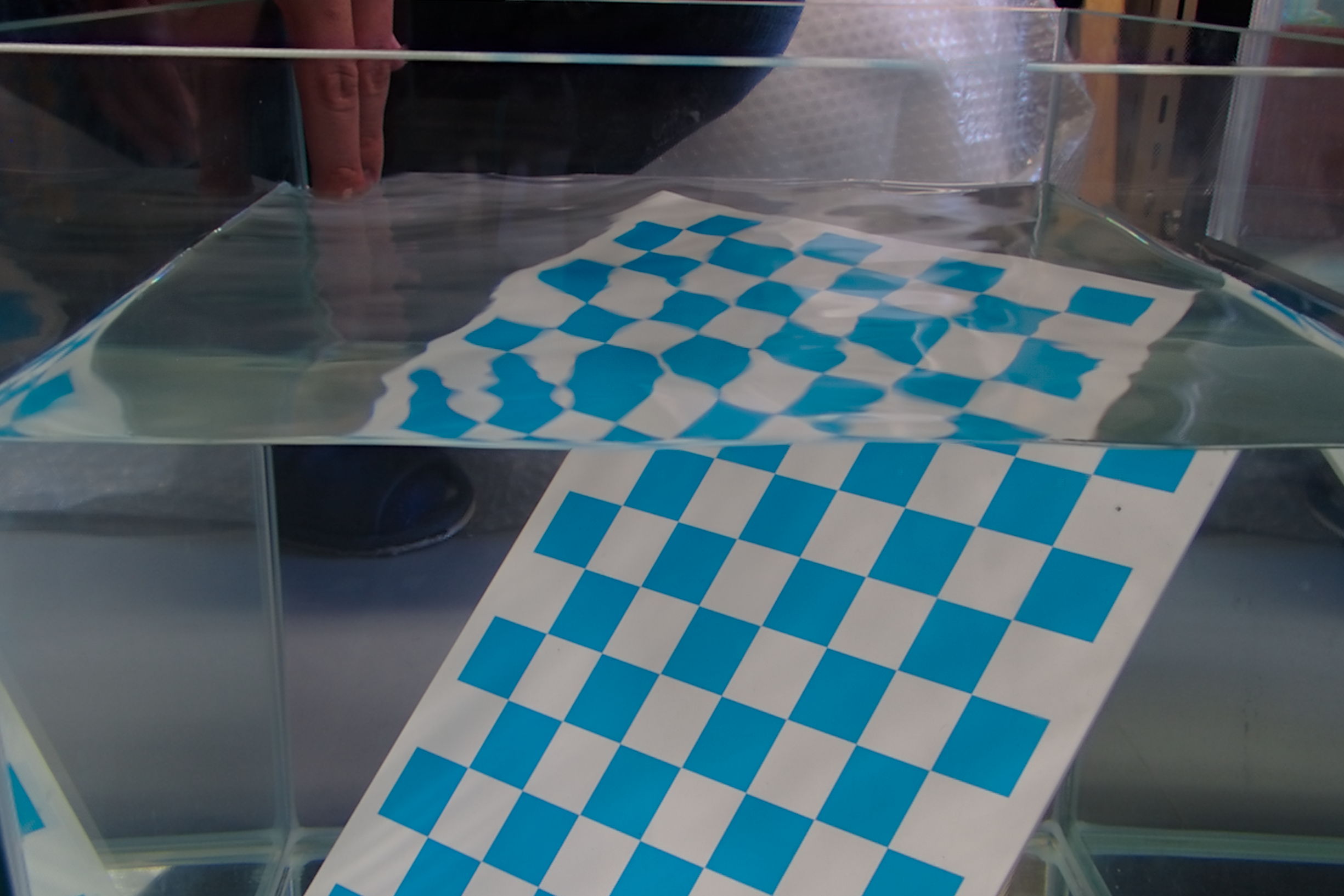}
    \end{minipage}
    \begin{minipage}{0.49\hsize}
        \centering
        \includegraphics[width=3.6cm]{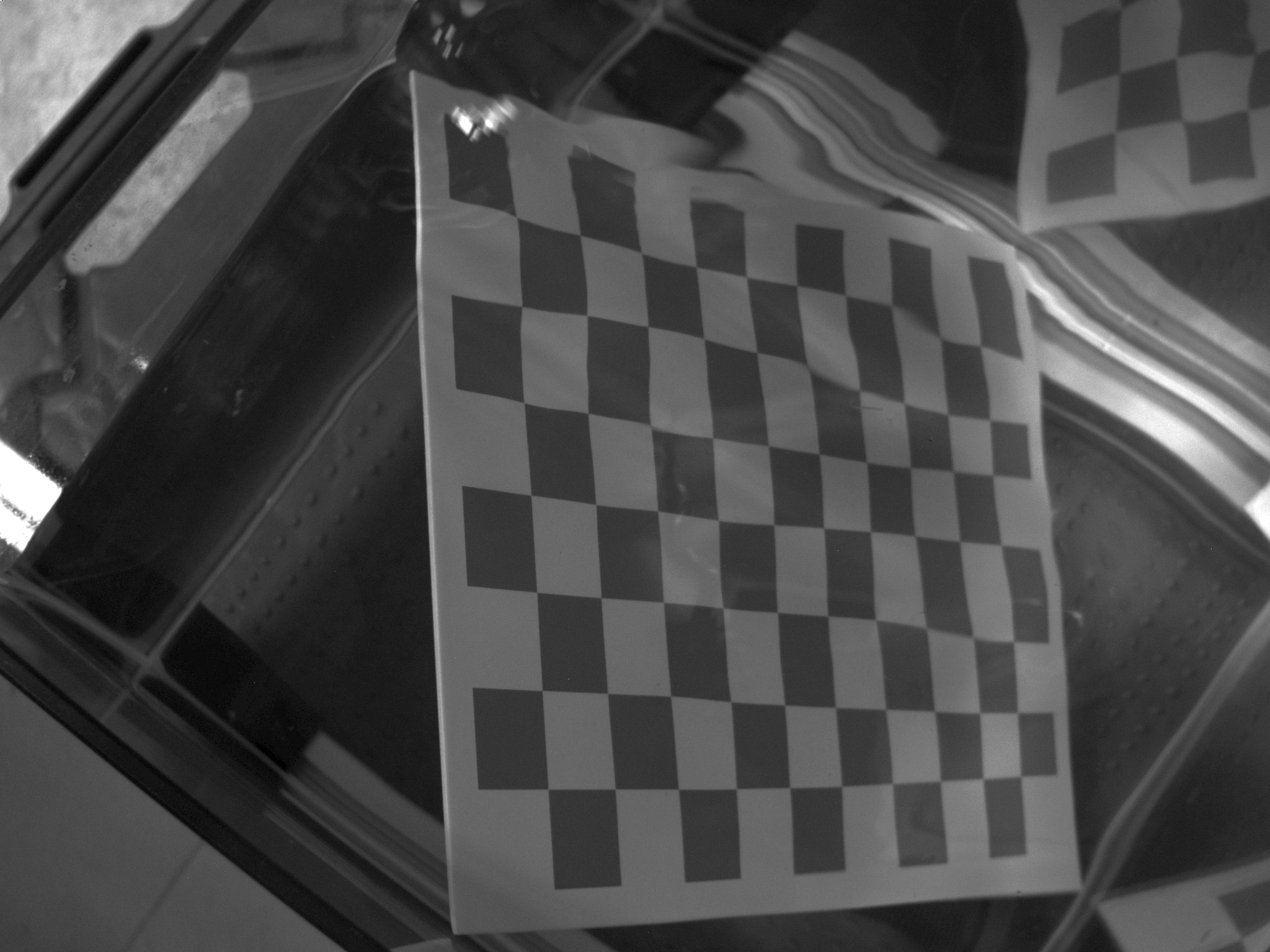}
    \end{minipage}
    \vspace{1mm}
    \caption{{\bf Left: }Appearance of the turbulence on the water surface. {\bf Right: }An example of the captured image.}
    \label{fig:wavy_water}
\end{figure}

\begin{figure}[t]
    \centering
    \includegraphics[width=7cm]{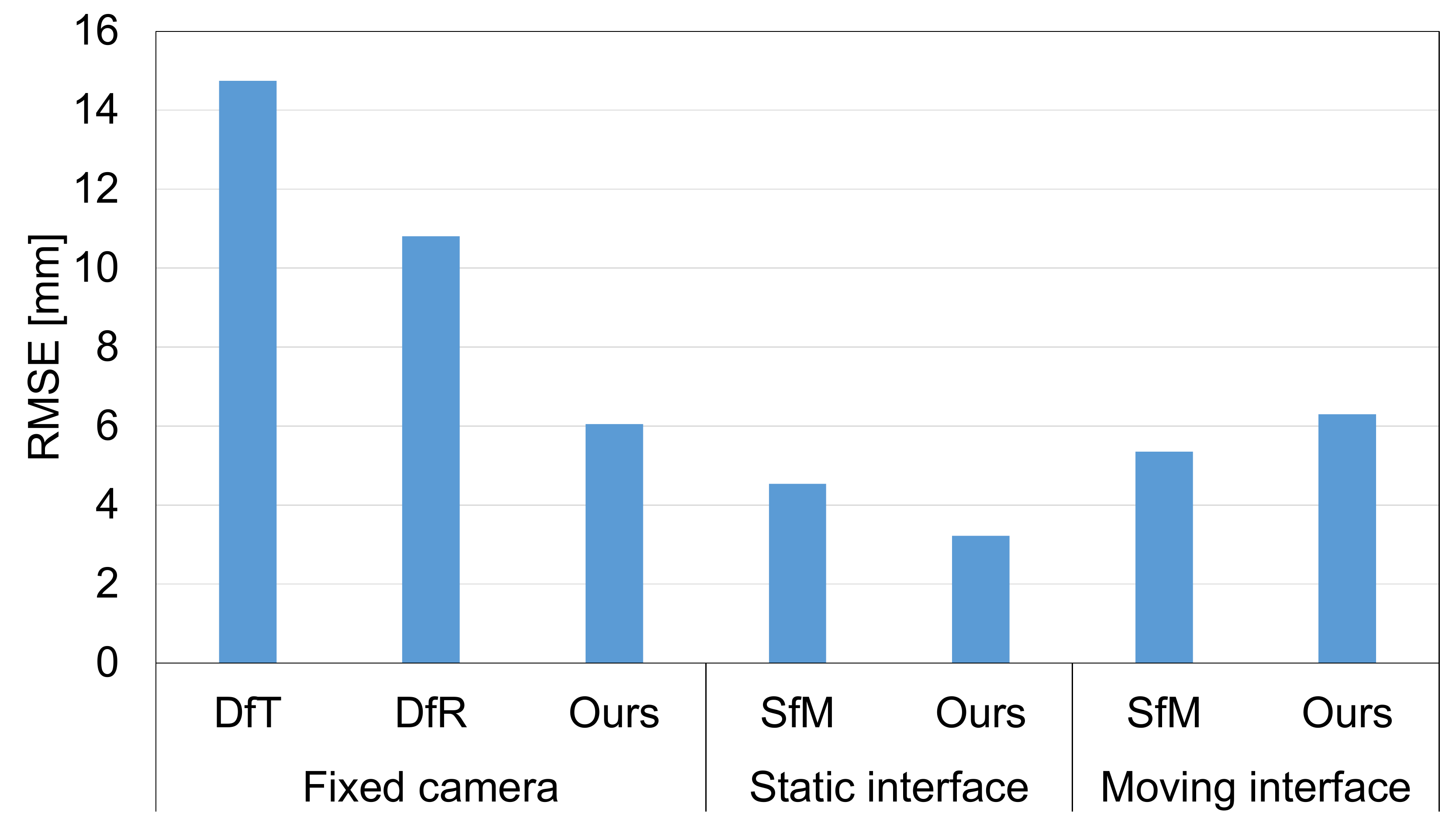}
    \caption{Quantitative result of the experiment with wavy refractive interface.}
    \label{fig:result_wave}
    \vspace{-0.5cm}
\end{figure}

Next, we evaluated the proposed method with a slightly wavy refractive interface, which is considered to be the condition where a water surface is disturbed by wind.
We placed a planar calibration board fixed on the bottom of the empty water tank, and the camera was set looking down on the water tank.
The board, the camera, and the tank moved together this time, and only water surface moved independently (\figref{expset}(right)).
We first captured images without water to obtain ground-truth and as a reference image for DfR.
Then, we poured water and captured an image for DfR while producing turbulence on the water surface by swinging hand.
Next, we inclined the water tank to several directions and angles, and captured a sequence of 20 images for each inclination while producing turbulence as fixed camera scenario.
We also captured 15 images with no inclination from various camera positions and directions as static interface scenario, and 20 images with several inclination and from various camera positions and directions as moving interface scenario, while producing turbulence.
\figref{wavy_water} shows an example of the turbulence on the water surface and captured image.

We reconstructed 3D points of the board using hard constraint and computed RMSE from the ground-truth shape as well as flat refractive interface scene.
The results are shown in \figref{result_wave}.
The proposed method performed better than the existing methods in fixed camera and static interface scenario.
We did not use explicit averaging as DfT, but we consider the effect of turbulence was intrinsically compensated by bundle adjustment.
In moving interface scenario, the proposed method performed slightly worse than simple SfM.
The proposed method successfully minimized the reprojection error more than simple SfM, but minimizing the reprojection error does not always lead to better 3D point reconstruction; the larger number of degrees-of-freedom in our geometric setup probably led to over-fitting in this specific case, where there was optical turbulence on the surface.

\subsection{Demo in fixed camera scenario}

\begin{figure}[t]
    \centering
    \begin{minipage}{0.49\hsize}
        \includegraphics[width=4cm]{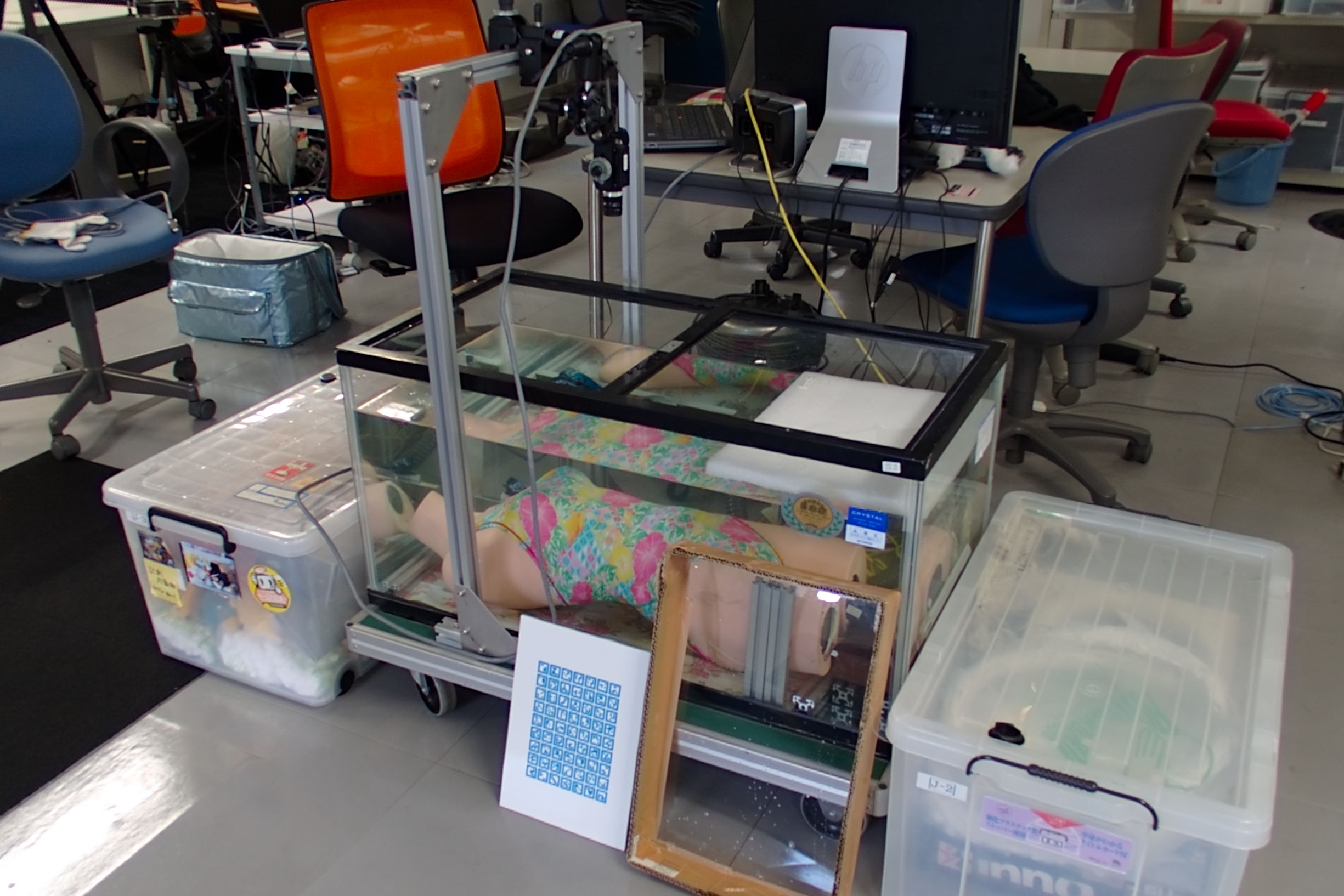}
    \end{minipage}
    \begin{minipage}{0.49\hsize}
        \includegraphics[width=4cm]{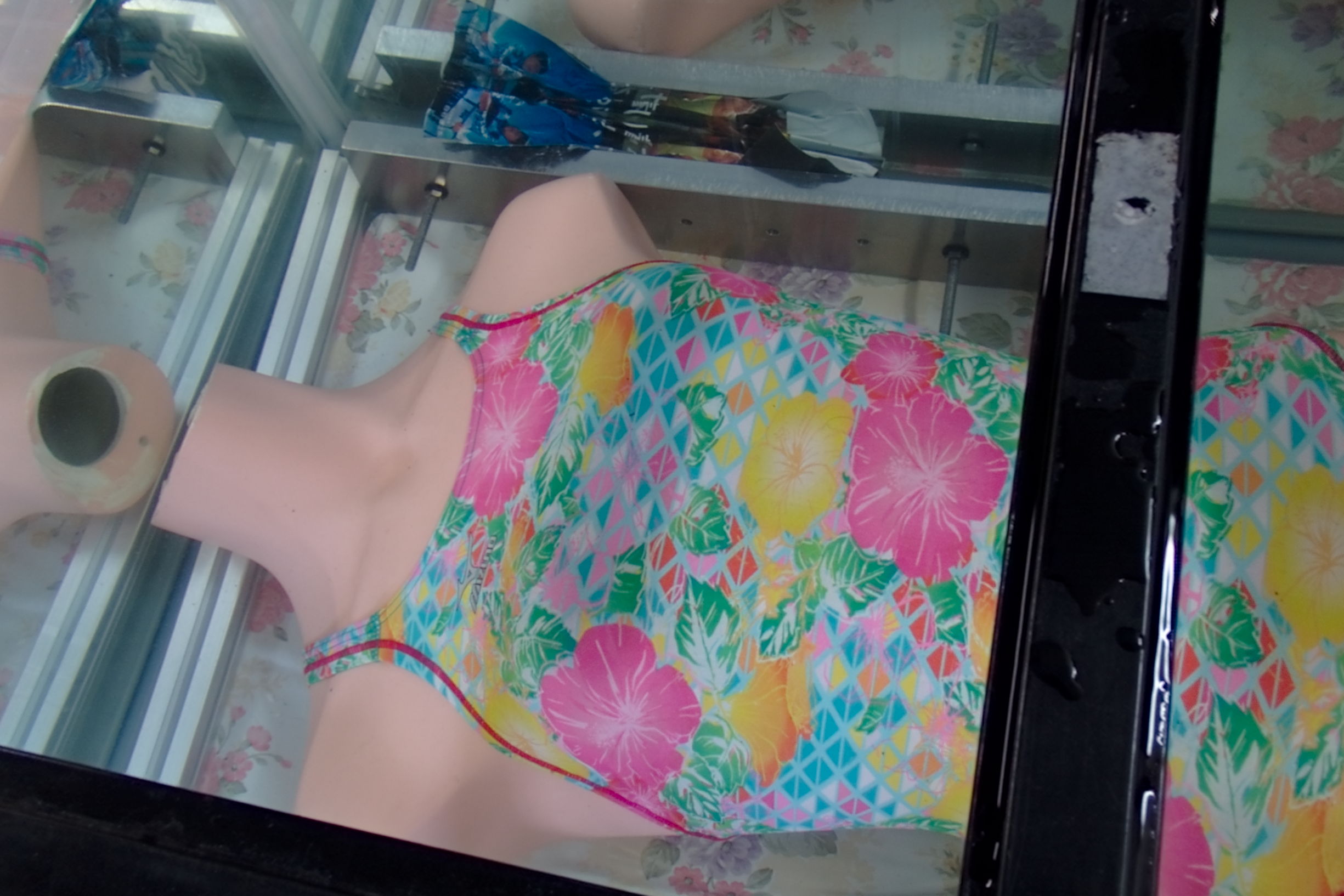}
    \end{minipage}
    \vspace{1mm}
    \caption{Experimental setup using a water surface as the refractive interface. In planar situation, a mannequin wearing swimsuit was submerged into the water tank.}
    \label{fig:expset2}
\end{figure}

\begin{figure}[t]
    \centering
    \begin{minipage}{0.49\hsize}
        \includegraphics[width=4cm]{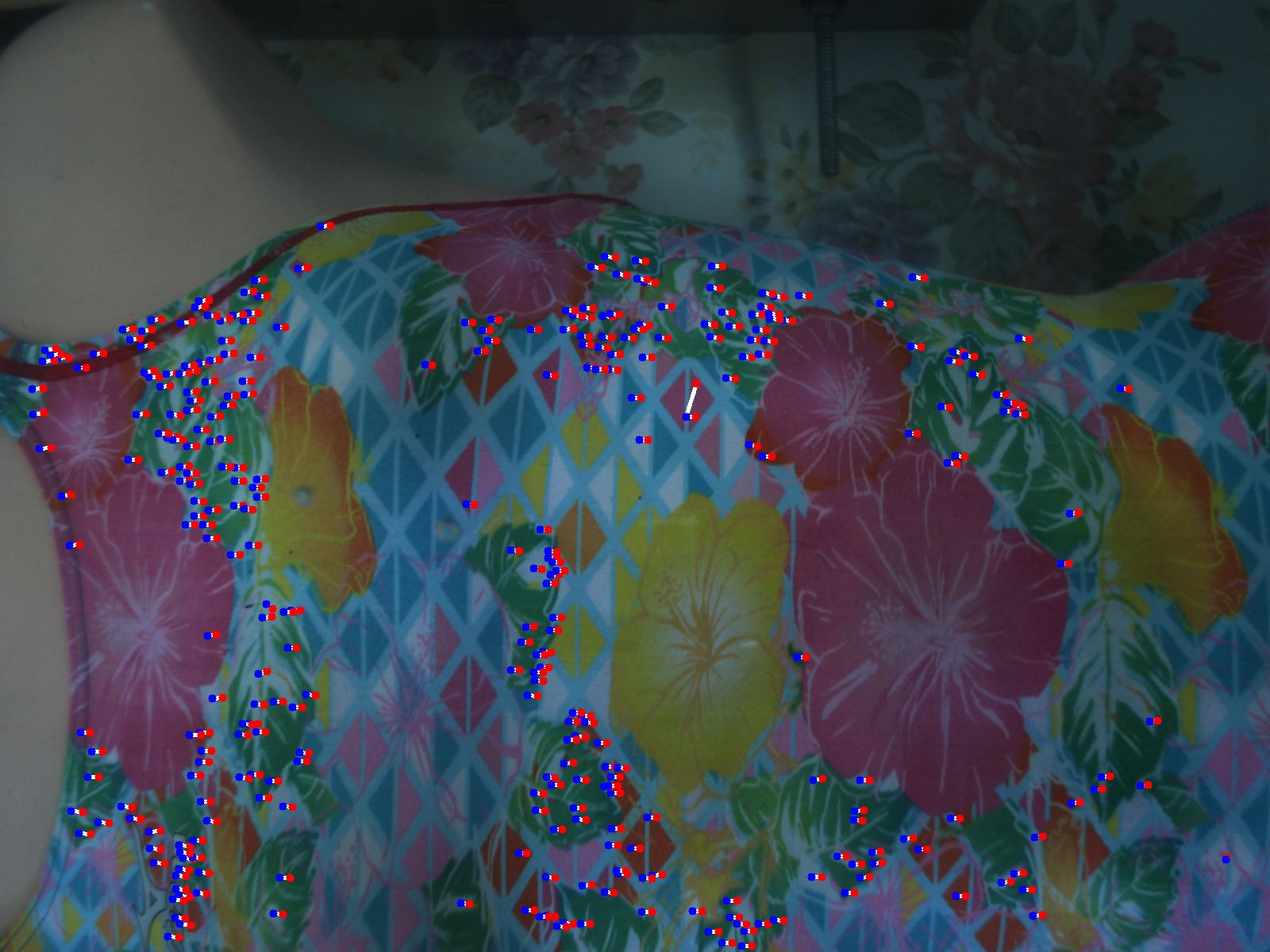}
    \end{minipage}
    \begin{minipage}{0.49\hsize}
        \includegraphics[width=4cm]{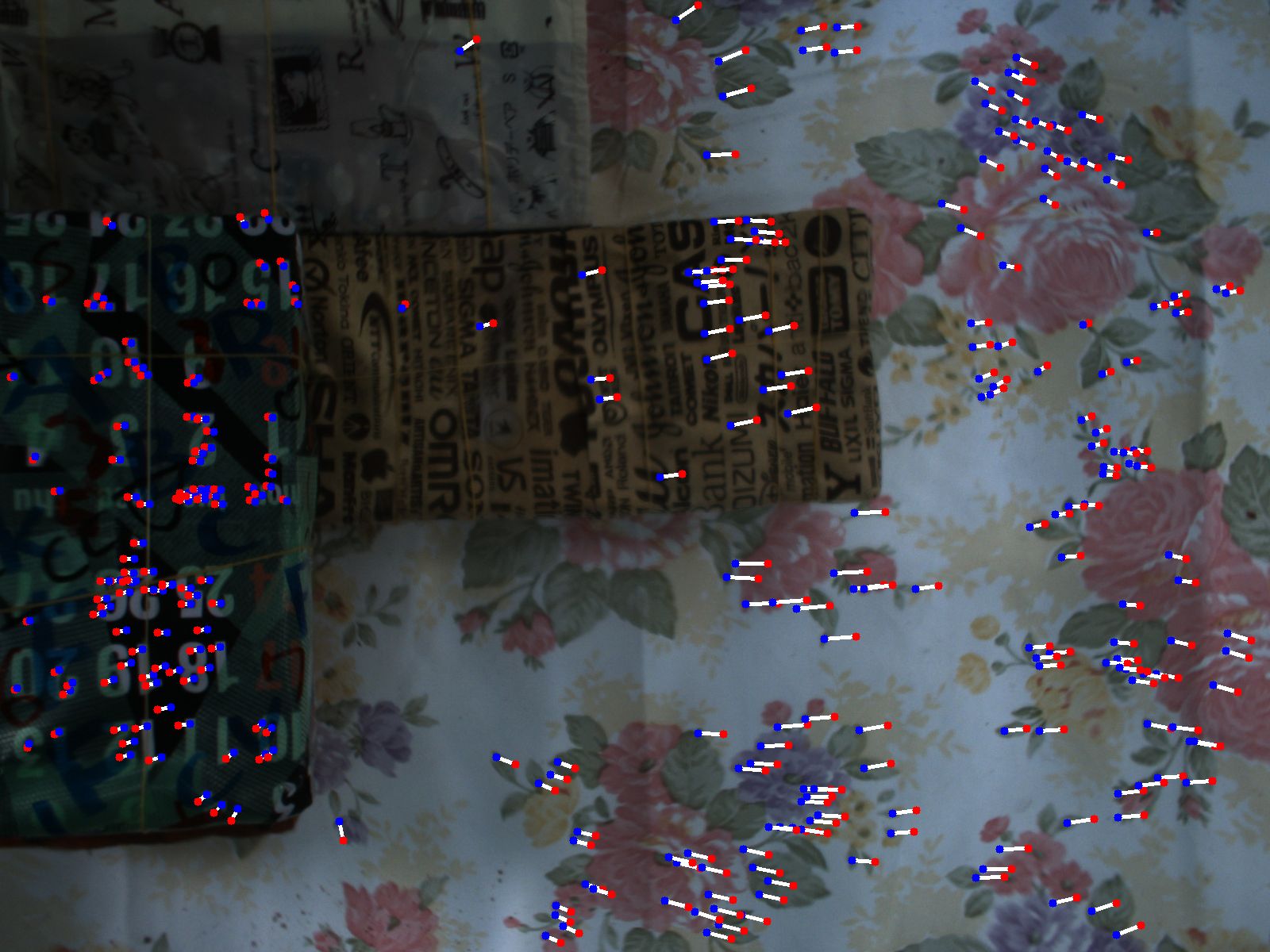}
    \end{minipage}
    \vspace{1mm}
    \caption{Examples of captured images in planar situation (left) and wavy situation (right). To emphasize the heavy turbulence, optical flow is visualized. }
    \label{fig:flow}
    \vspace{-0.5cm}
\end{figure}

In order to confirm the applicability of the proposed method in real-world underwater scenes, we also conducted qualitative evaluations under several conditions.
First, we evaluated the proposed technique in fixed camera scenario.
We tried two situations: in one situation a transparent plane was placed on the water surface and moved to make the surface planar (planar situation); in the other situation we slowly moved the water tank up and down to incline the water surface (wavy situation).
In both situations, we used a water tank of $90\times45\times45cm$ and a camera Point Grey Grasshopper 2.
The camera was fixed on the water tank, \ie, the camera also moved when the water tank moved.
\figref{expset2} shows the experimental setup.
Examples of captured images are shown in \figref{flow}, showing that turbulence is observed in wavy situation.

\begin{figure}[t]
    \centering
    \begin{minipage}{0.49\hsize}
        \includegraphics[width=4.0cm]{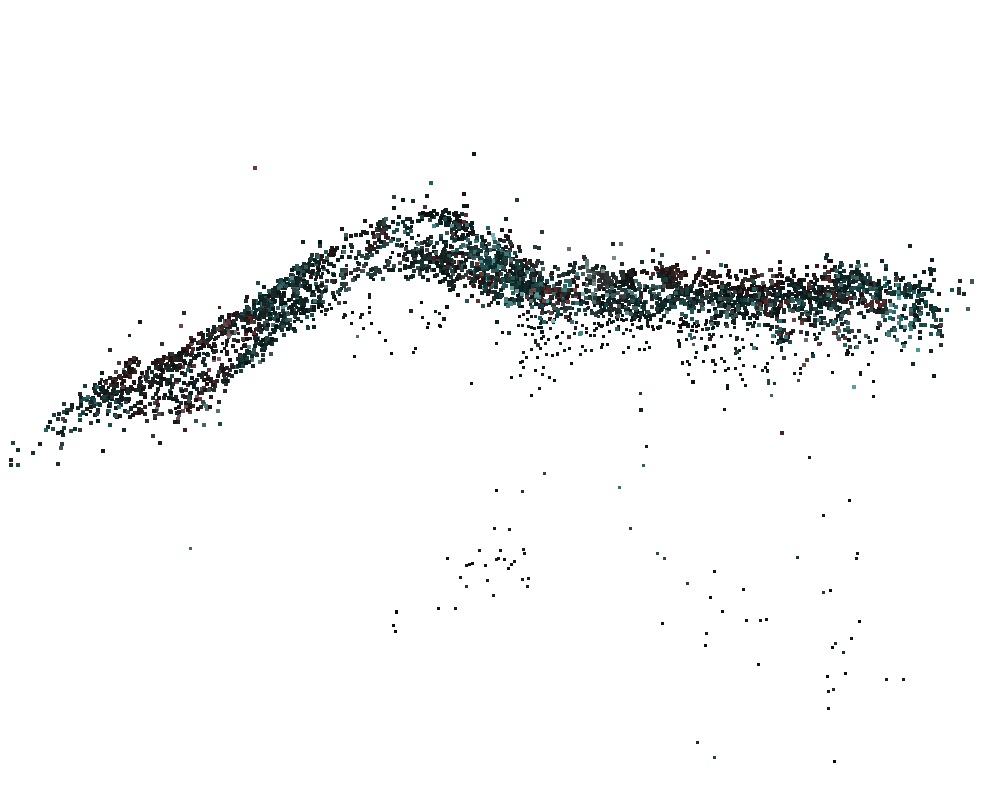}
    \end{minipage}
    \begin{minipage}{0.49\hsize}
        \includegraphics[width=4.0cm]{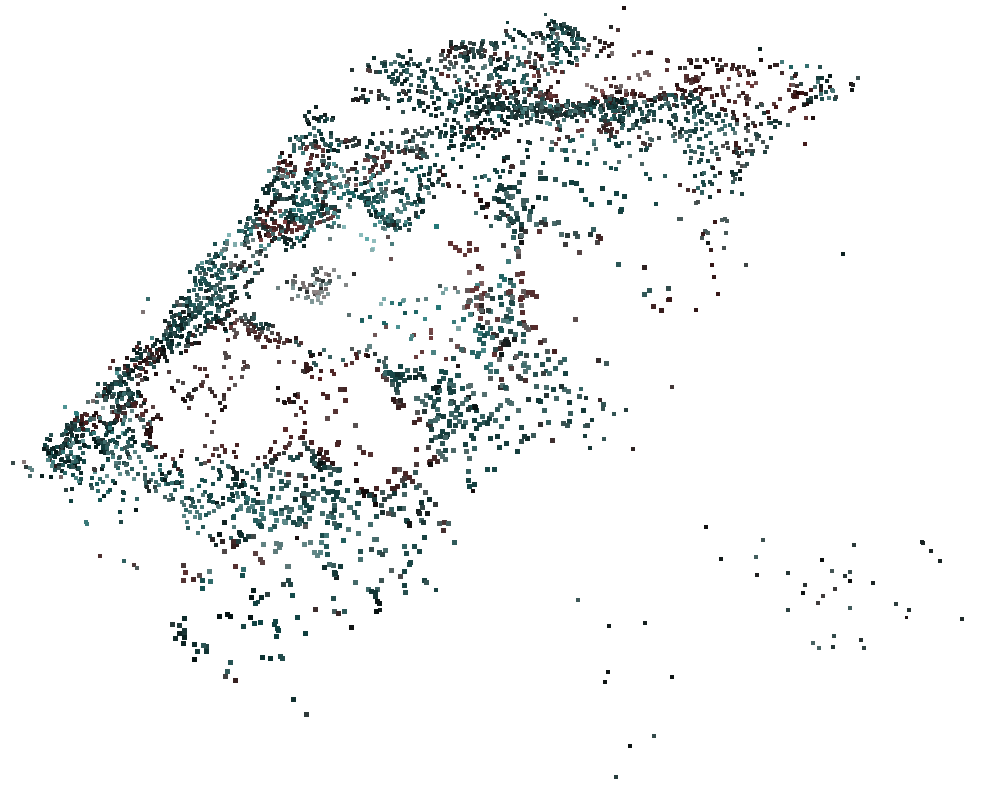}
    \end{minipage}
    \vspace{1mm}
    \caption{Reconstructed point clouds of the captured mannequin.}
    \label{fig:mannequin_pc}
\end{figure}

\begin{figure}[t]
    \centering
    \begin{minipage}{0.32\hsize}
        \includegraphics[width=2.5cm]{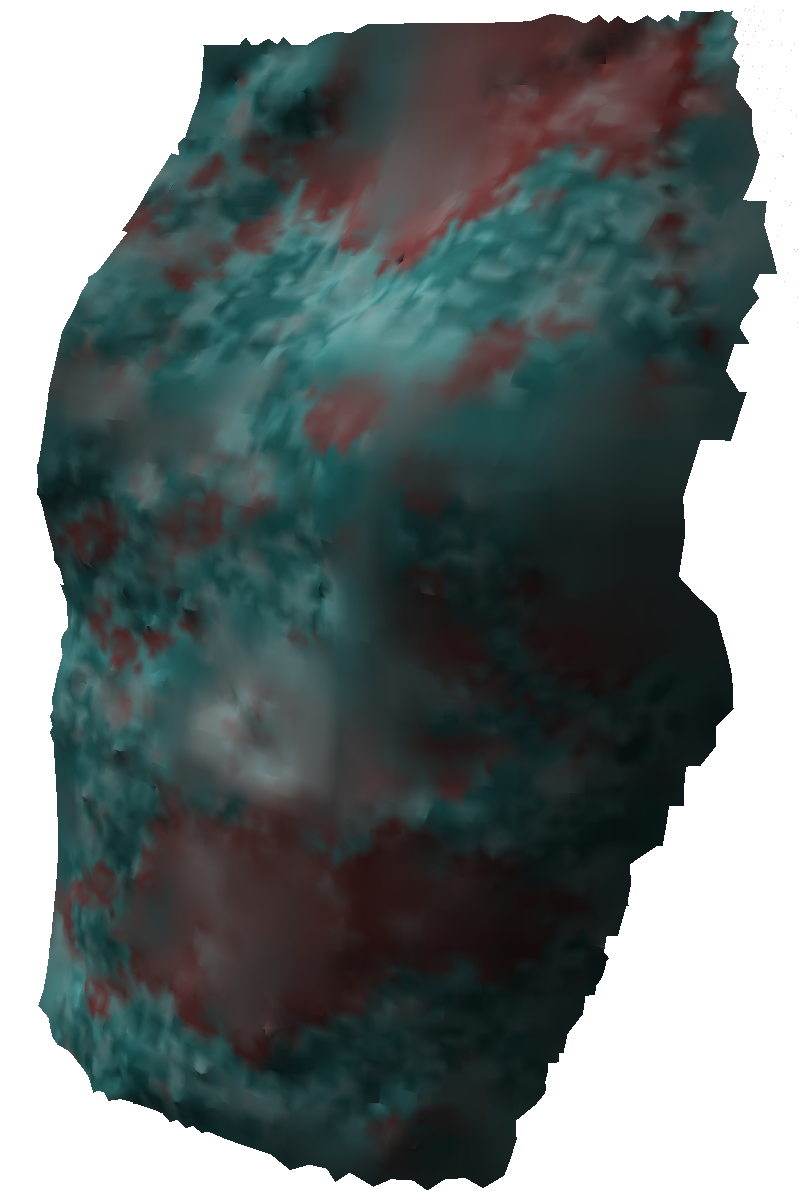}
    \end{minipage}
    \begin{minipage}{0.32\hsize}
        \includegraphics[width=2.5cm]{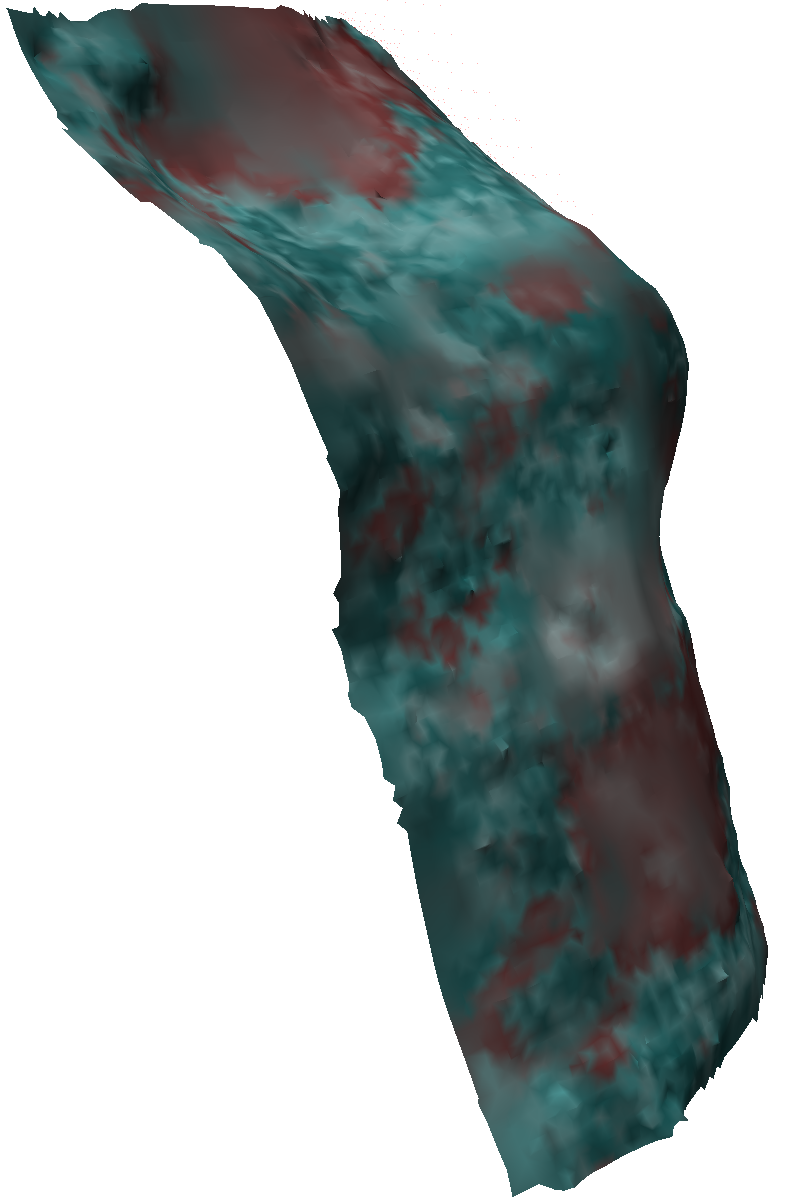}
    \end{minipage}
    \begin{minipage}{0.32\hsize}
        \includegraphics[width=2.5cm]{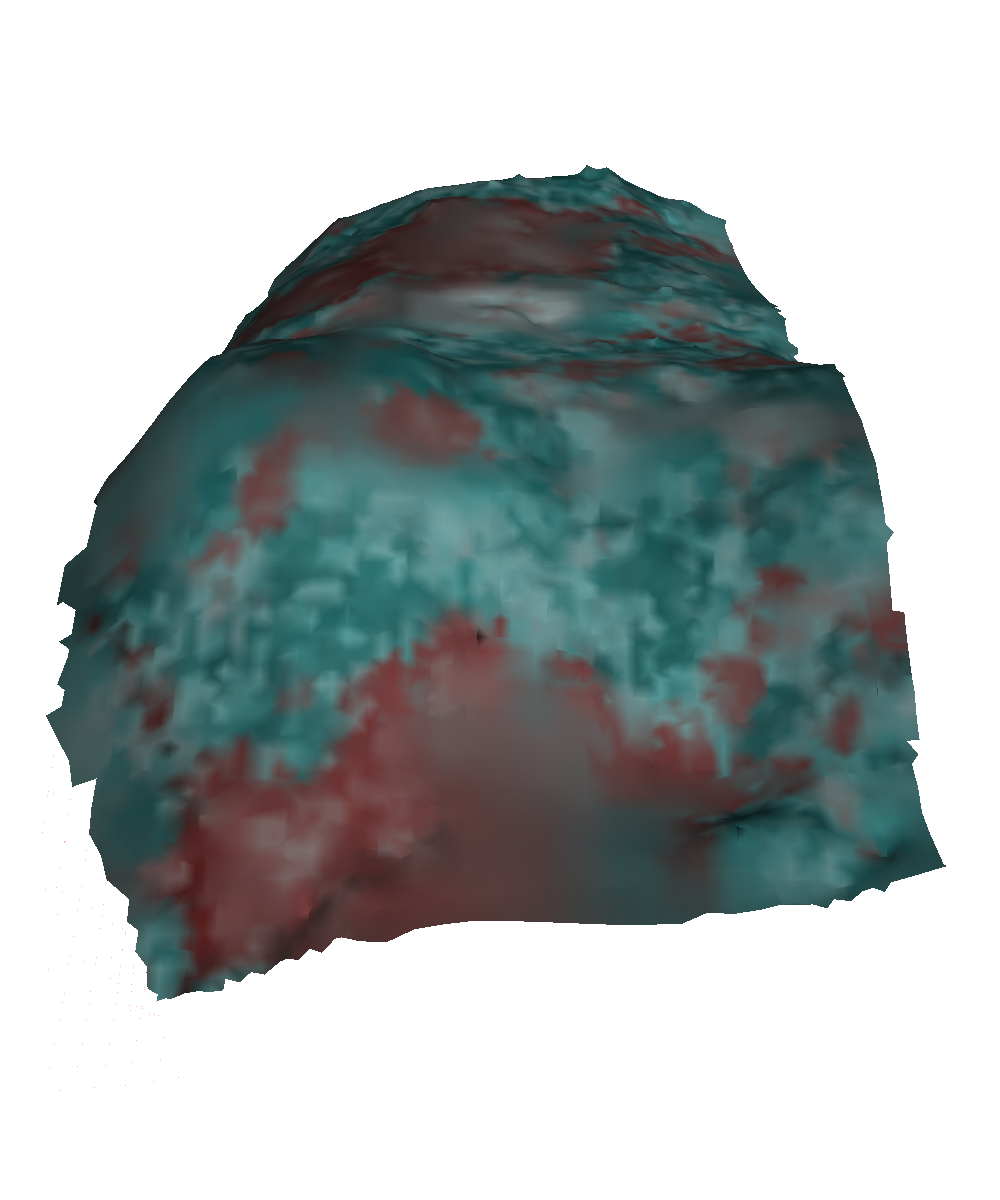}
    \end{minipage}
    \begin{minipage}{0.32\hsize}
        \includegraphics[width=2.5cm]{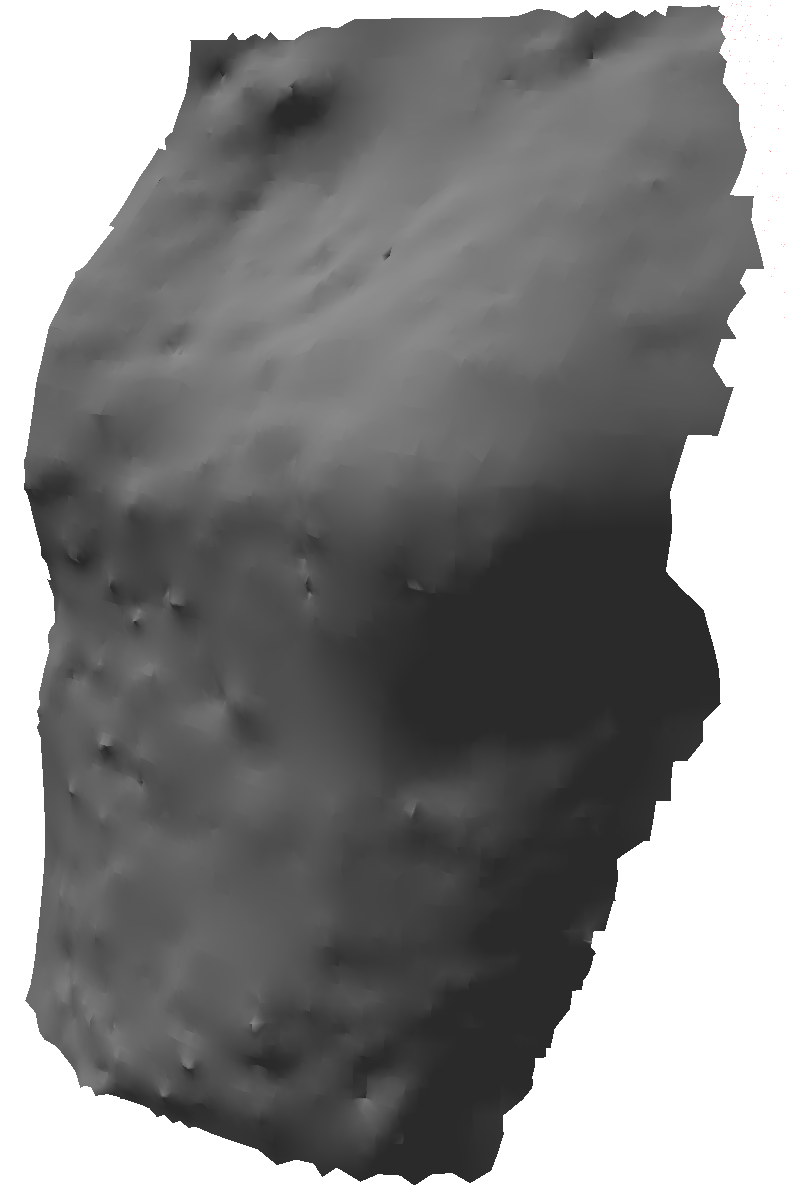}
    \end{minipage}
    \begin{minipage}{0.32\hsize}
        \includegraphics[width=2.5cm]{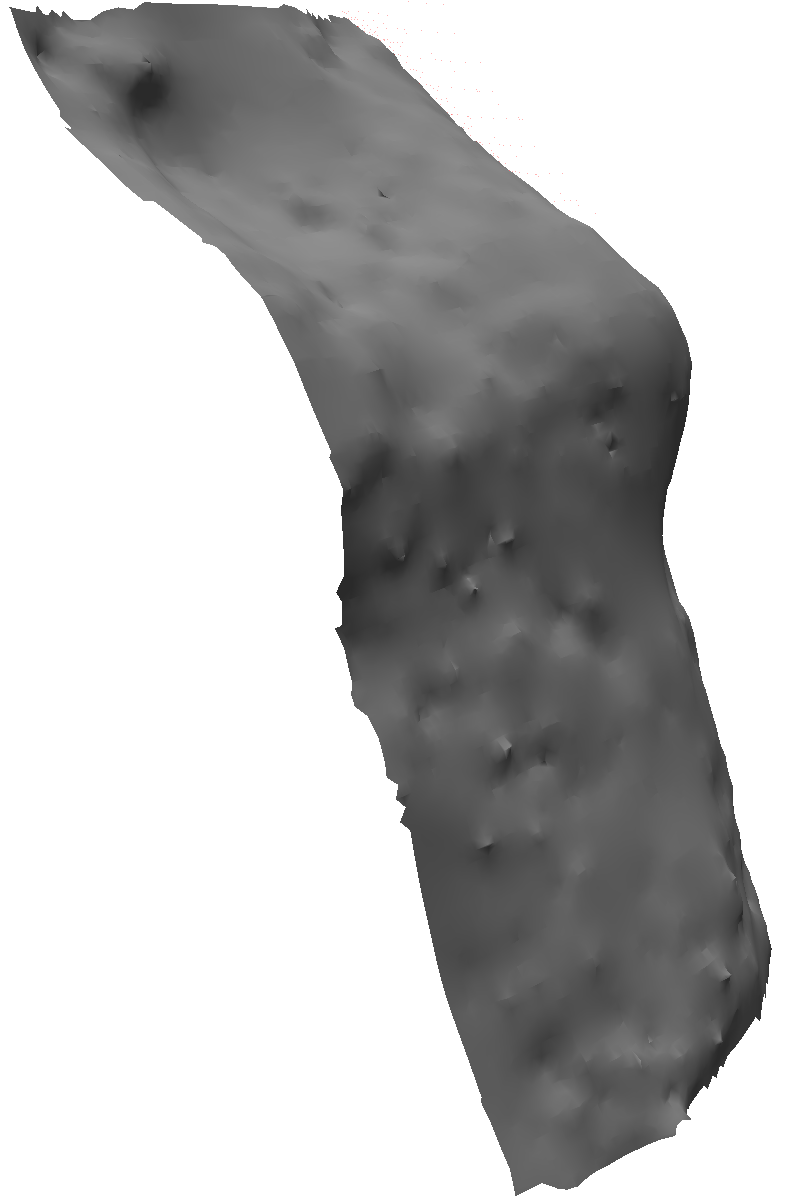}
    \end{minipage}
    \begin{minipage}{0.32\hsize}
        \includegraphics[width=2.5cm]{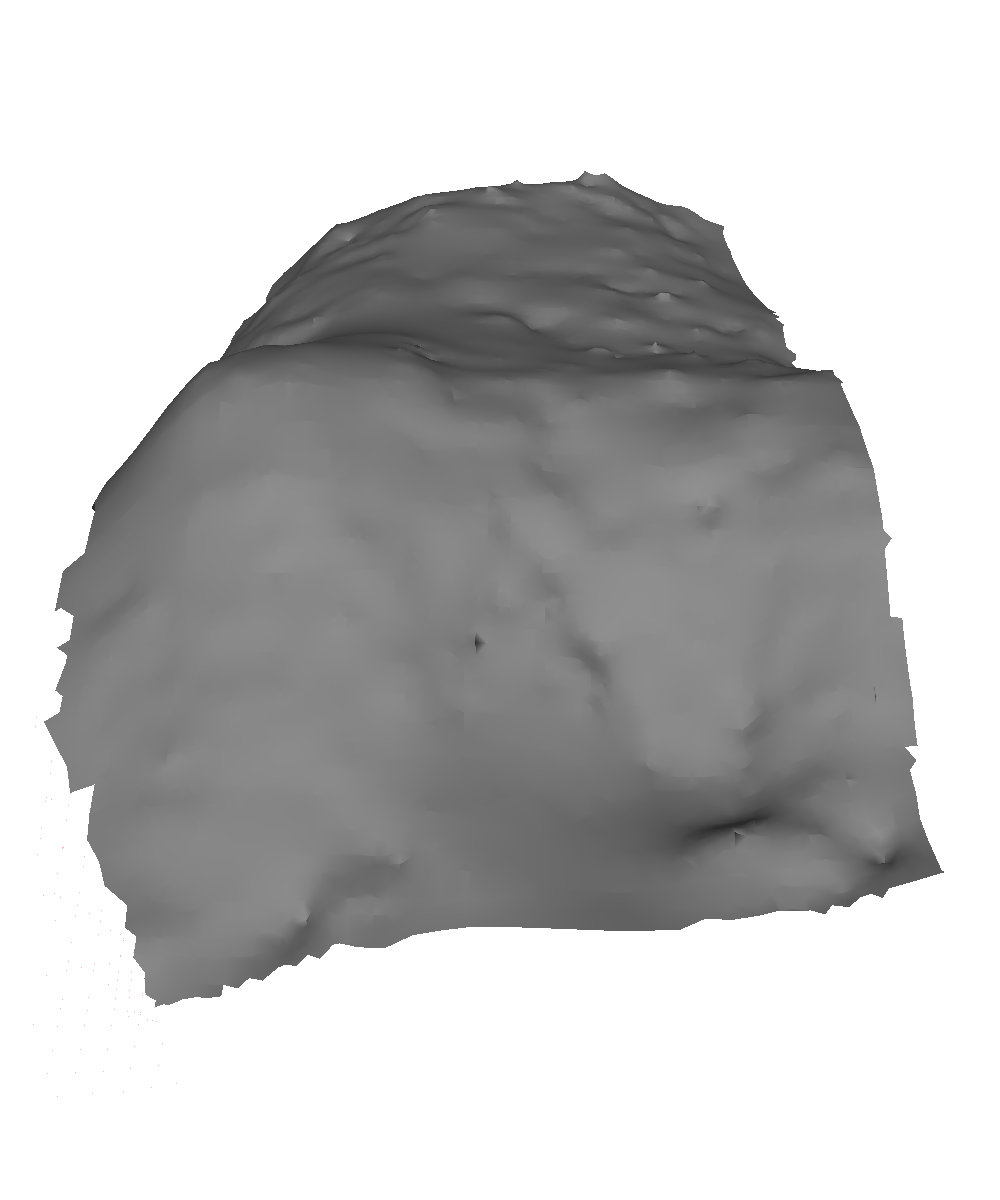}
    \end{minipage}
    \vspace{1mm}
    \caption{Reconstructed mesh of the captured mannequin. Mesh was reconstructed from point clouds by Poisson equation~\cite{Kazhdan:EGSGP06}. Upper row images show the textured model from several angles, and lower row images shows shading.}
    \label{fig:mannequin}
\end{figure}

\begin{figure}[t]
    \centering
    \begin{minipage}{0.49\hsize}
        \includegraphics[width=4cm]{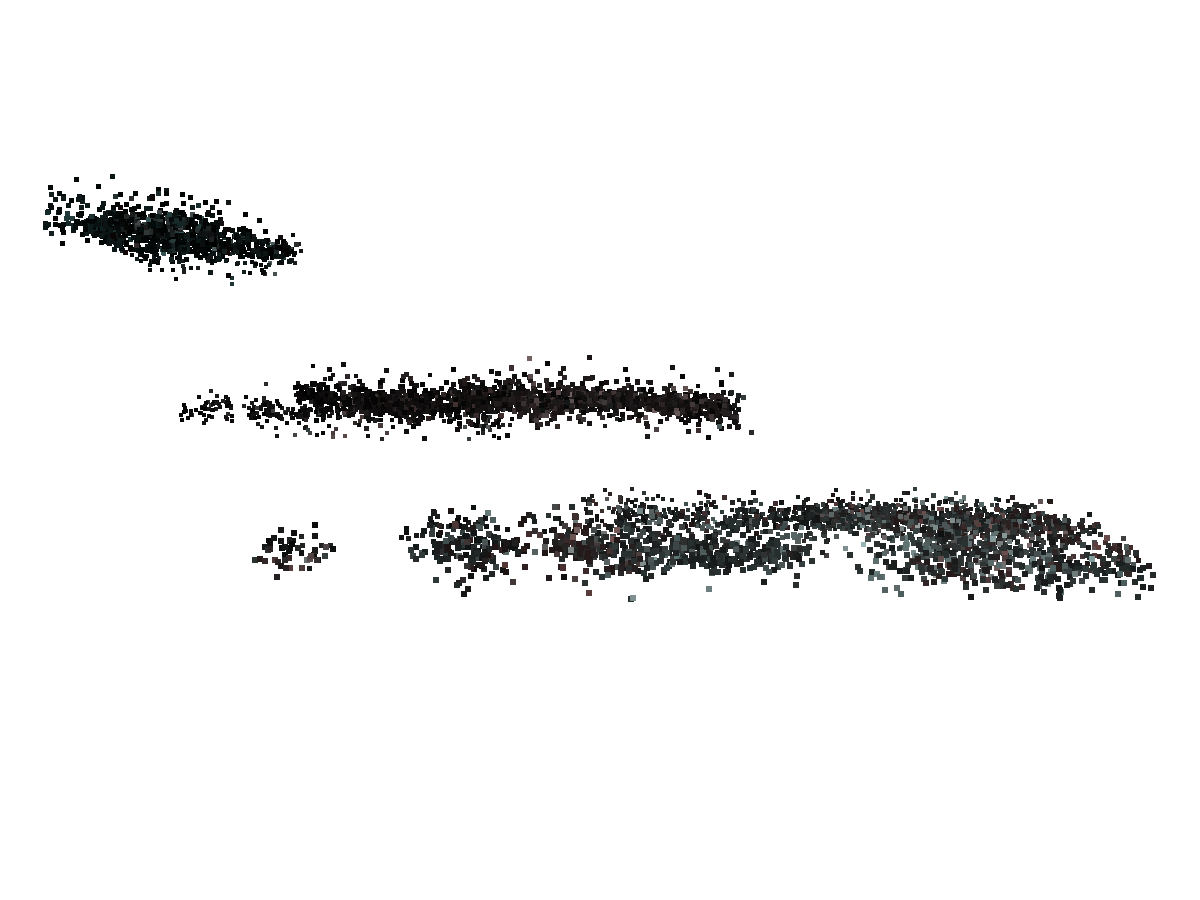}
    \end{minipage}
    \begin{minipage}{0.49\hsize}
        \includegraphics[width=4cm]{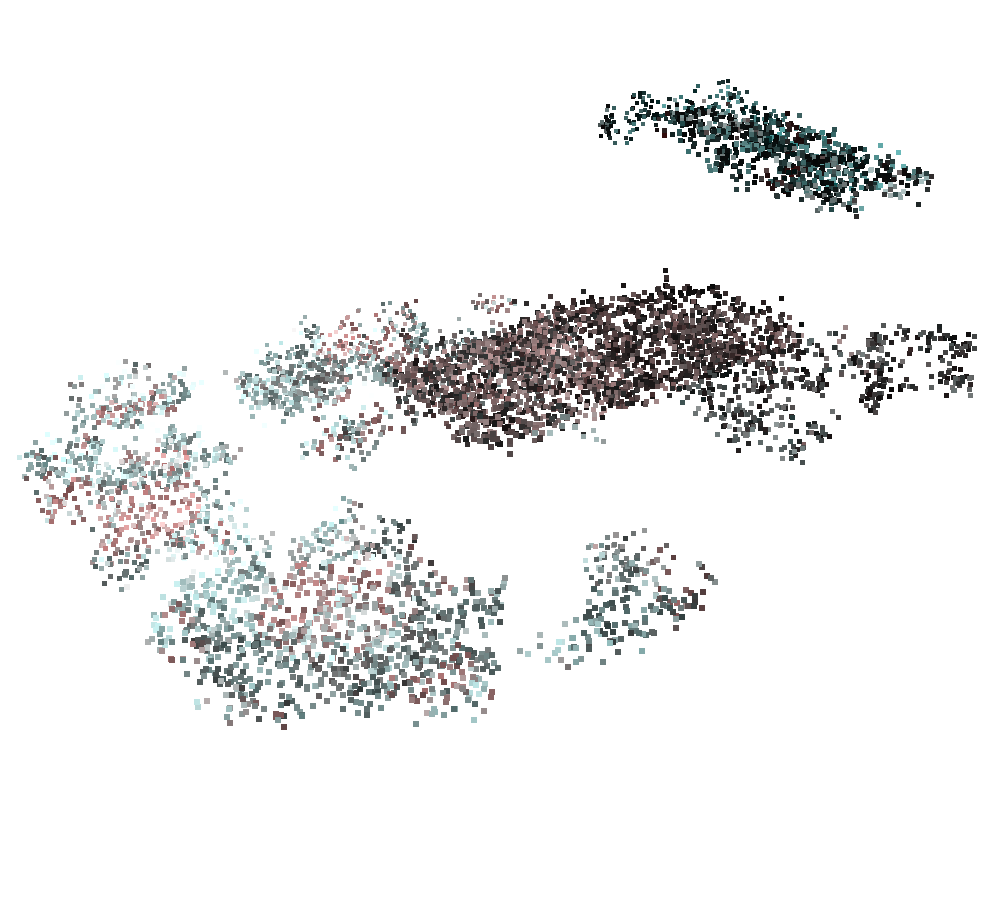}
    \end{minipage}
    \caption{Reconstructed point clouds of the captured brick under wavy water surface. }
    \label{fig:brick}
    \vspace{-0.5cm}
\end{figure}

In planar situation, we used a mannequin as the measurement target.
We fixed it on the bottom of the water tank, and placed the camera above the water surface.
We placed a transparent acrylic plate on the water surface so that the camera would not observe turbulence due to small waves.
Then, we moved the water surface to make a big wave, and the acrylic plate tilted significantly.
We captured 240 images, and selected 26 images and a reference image according to the distance of optical flow.
Finally we applied our hard constraint method, which produced qualitatively pleasing reconstruction result as shown in \figref{mannequin_pc} and \figref{mannequin}.

In wavy situation, we used bricks wrapped in the wrapping sheet with much texture.
We also fixed them on the bottom of the water tank, and placed the camera above the water surface.
Then we slowly moved the water tank up and down to incline the water surface.
We captured 100 images per each inclination and 10 types of inclination, \ie, 1000 images in total.
This time, we suffered from heavy turbulence due to small waves (\figref{flow}).
In order to compensate for such effects, we averaged 2D point coordinates along each inclination to obtain 10 averaged flow data, which was used as the input to our hard constraint method.
Note that the proposed method worked well without averaging, but explicit averaging improved the quality, by exploiting the fact that multiple images shared the same refractive interface parameters.
The reconstruction result is shown in \figref{brick}.
The shape was accurately reconstructed despite the images were captured in heavy turbulence.

\subsection{Demo in static and moving interface scenario}
\label{ssec:demo_sfm}
\begin{figure}[t]
    \centering
    \begin{minipage}{0.49\hsize}
        \centering
        \includegraphics[width=4cm]{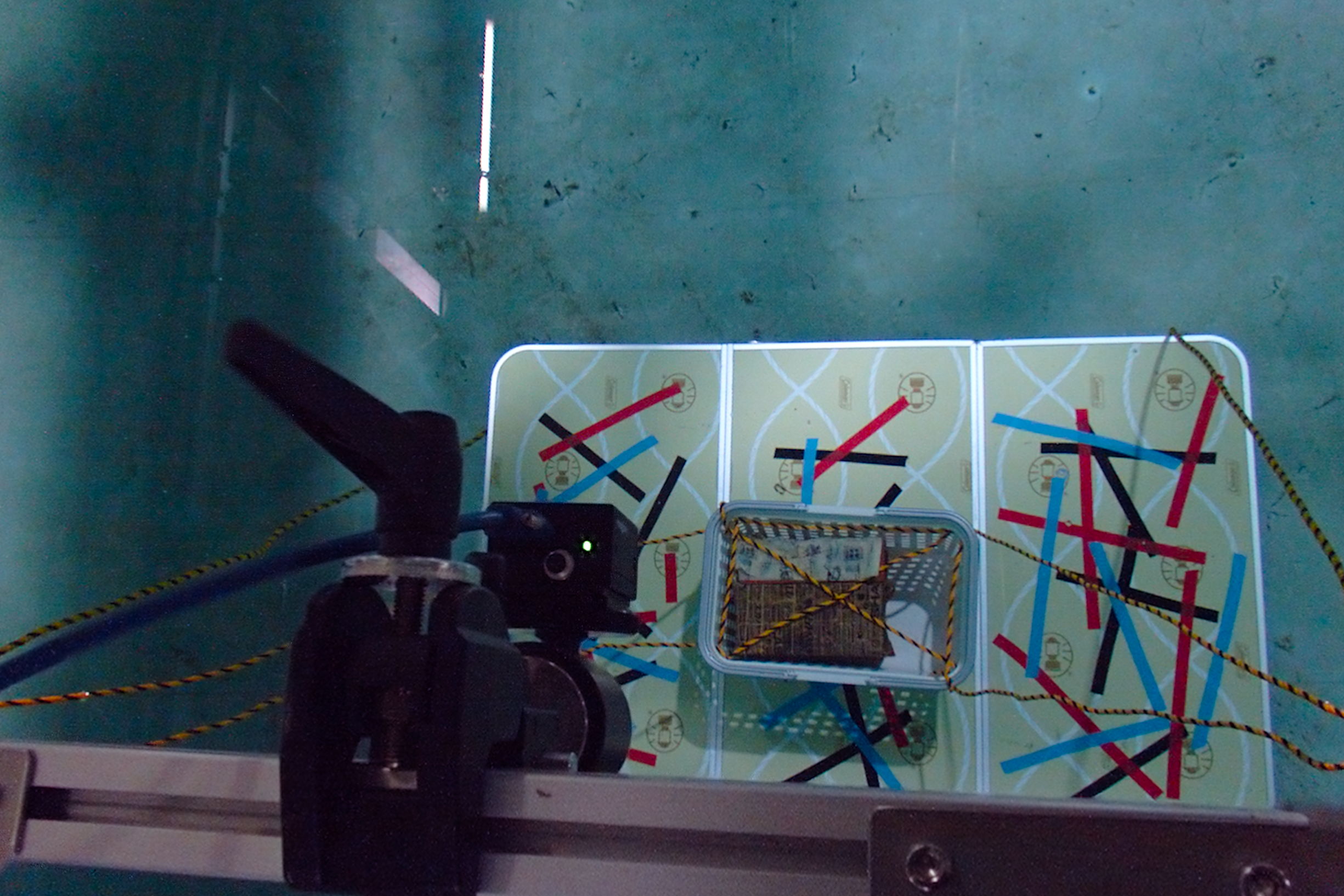}
    \end{minipage}
    \begin{minipage}{0.49\hsize}
        \centering
        \includegraphics[width=3.6cm]{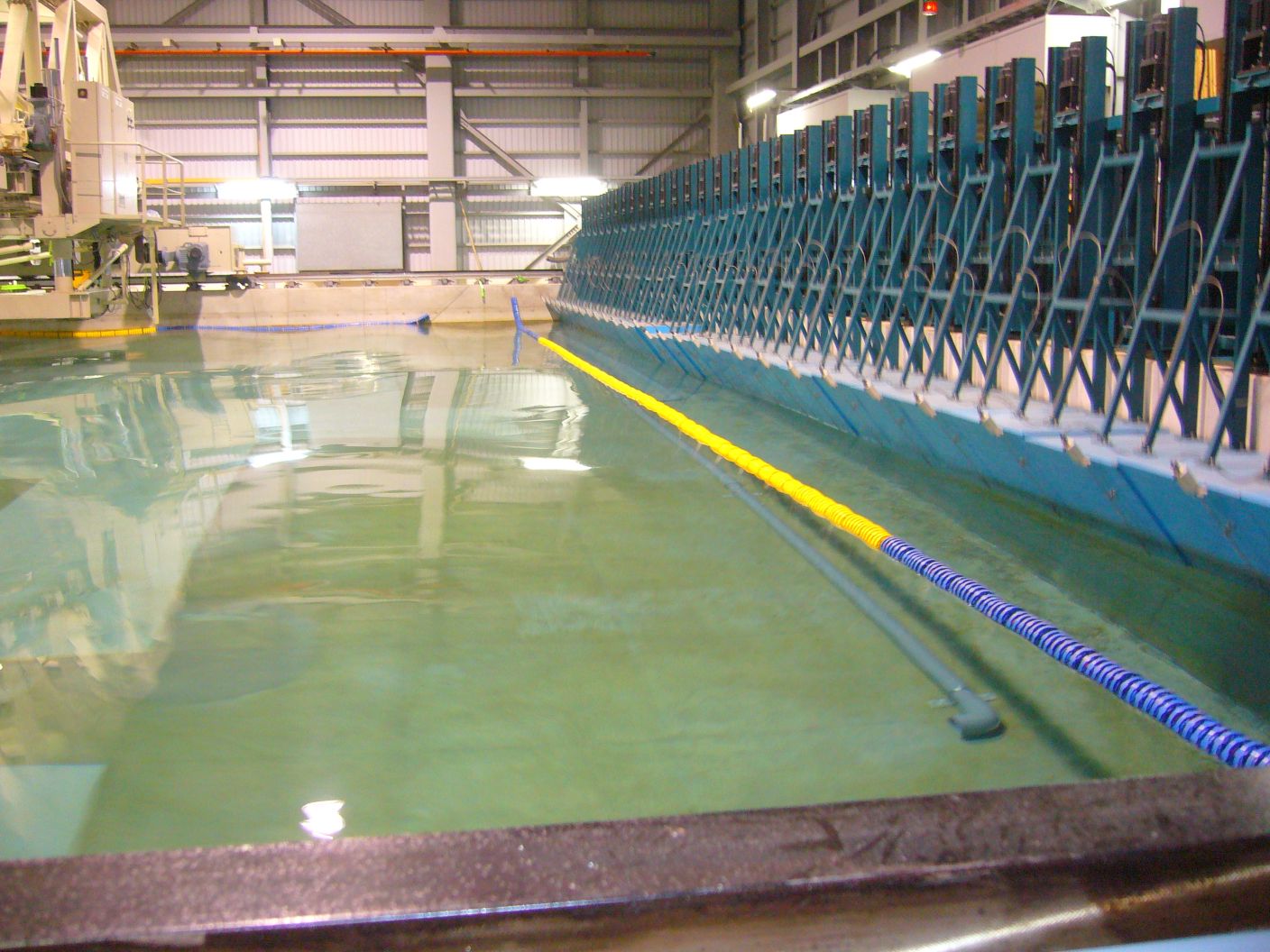}
    \end{minipage}
    \vspace{1mm}
    \caption{Appearance of the experimental setup in \subsecref{demo_sfm}.}
    \label{fig:expset3}
    \vspace{-0.5cm}
\end{figure}

Next, we evaluated the proposed technique in static and moving interface scenario.
We conducted the experiments in the pool with a wave generator.
The pool was filled with transparent water of $1.8m$ height from the bottom.
As measurement targets, we submerged a table of $60cm$ height, and placed a basket of $30cm$ height with bricks on the table, as shown in \figref{expset3}(left).
The camera was placed on the rig $60cm$ above the water surface.
First, we captured 20 images of the targets in the still water while moving the camera along a line as static interface scenario.
Second, we generated waves by the wave generator to make $1$ second period wave, and the water surface tilted as shown in \figref{expset3}(right).
Then, we captured 20 images by the moving camera as moving interface scenario.

We applied the proposed method respectively to reconstruct the targets.
We considered generated waves are almost planar, because the wavelength was much longer than the camera's field of view.
However, the wave was not completely planar, and had slight curve, where soft constraint with local regularizer is expected to work well.
Thus we tried it when we reconstruct moving interface scenes.

\begin{figure}[t]
    \centering
    \begin{minipage}{0.49\hsize}
        \includegraphics[width=4cm]{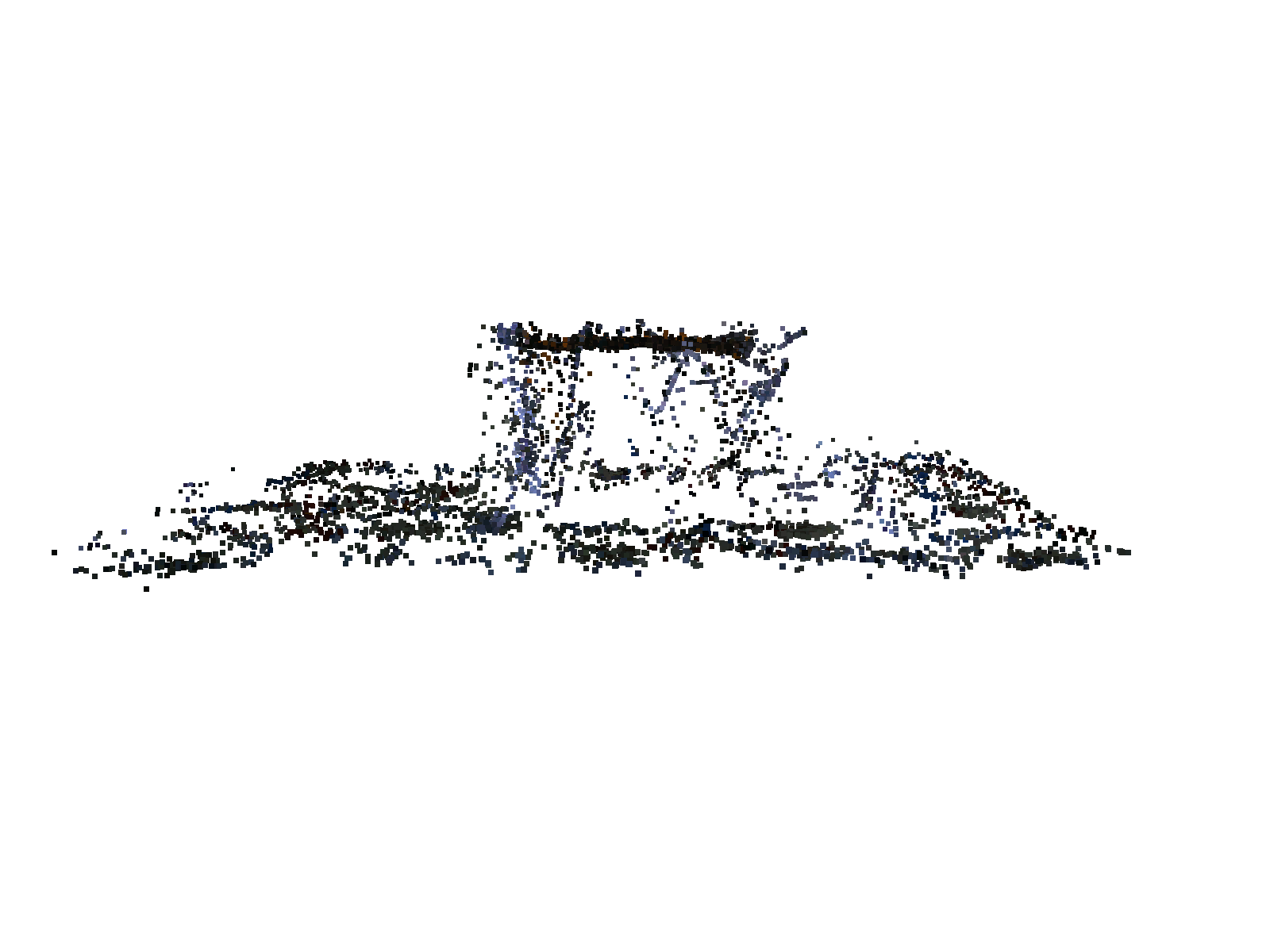}
    \end{minipage}
    \begin{minipage}{0.49\hsize}
        \includegraphics[width=4cm]{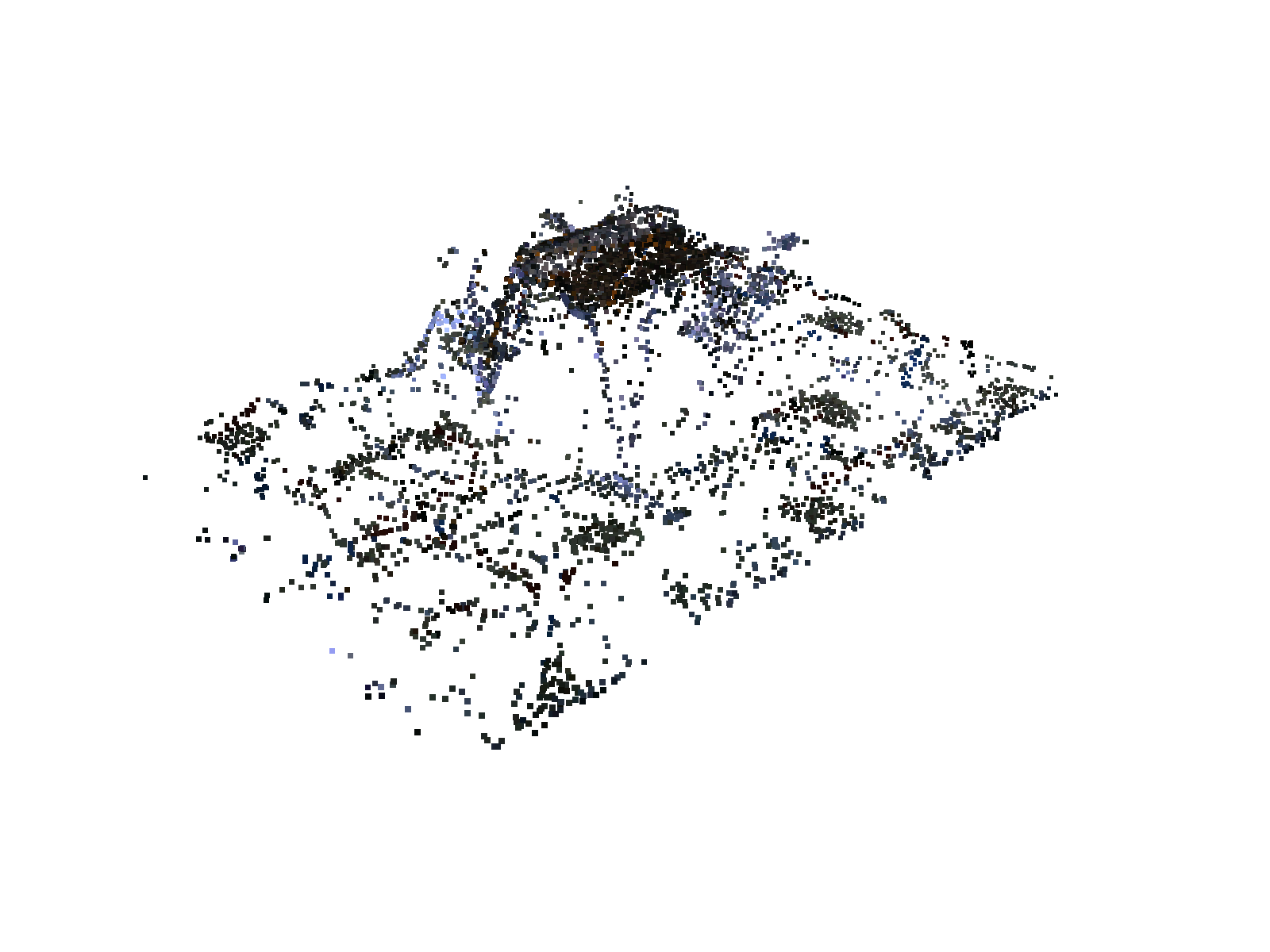}
    \end{minipage}
    (a) Static interface with hard constraint\\
    \begin{minipage}{0.49\hsize}
        \includegraphics[width=4cm]{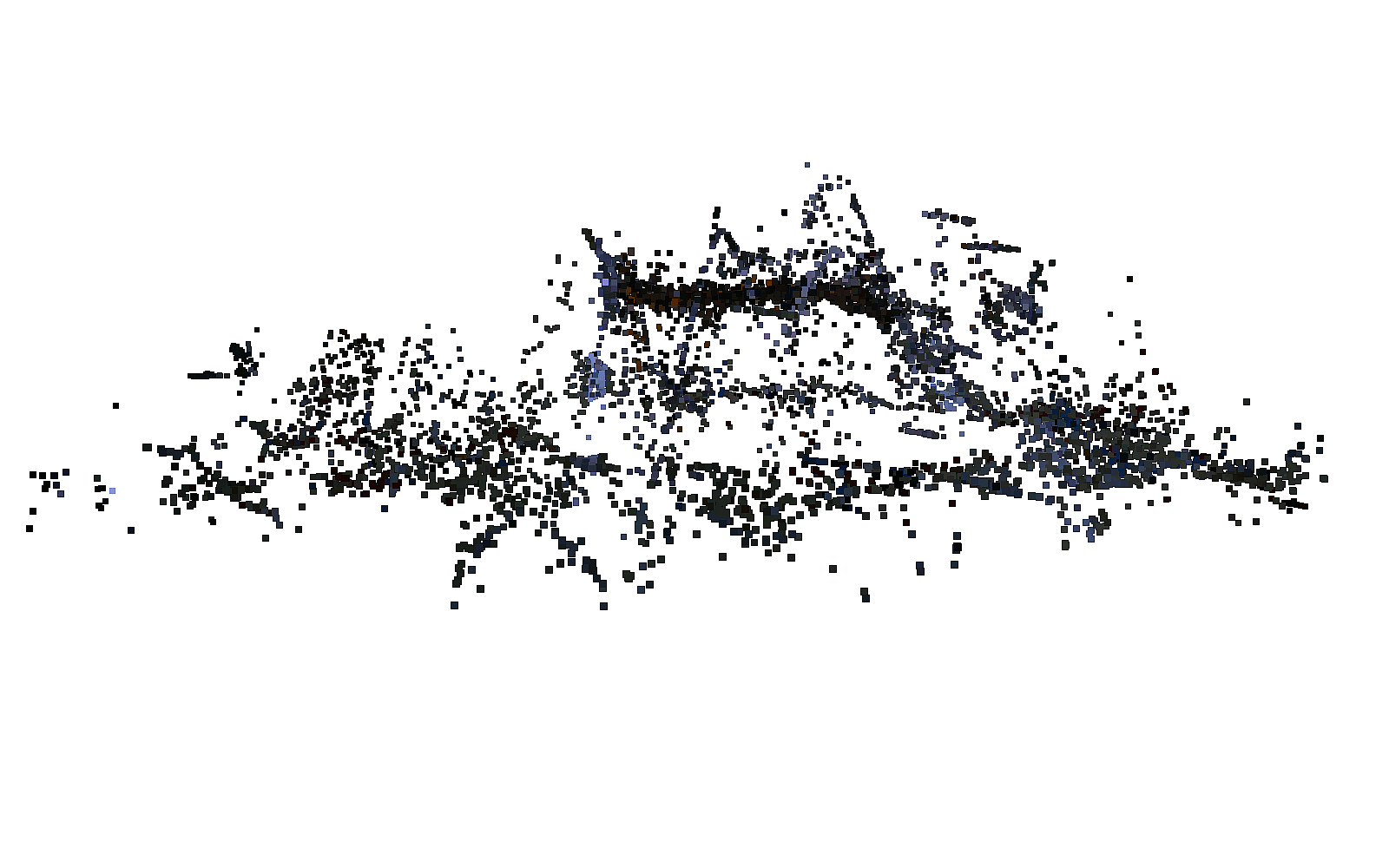}
    \end{minipage}
    \begin{minipage}{0.49\hsize}
        \includegraphics[width=4cm]{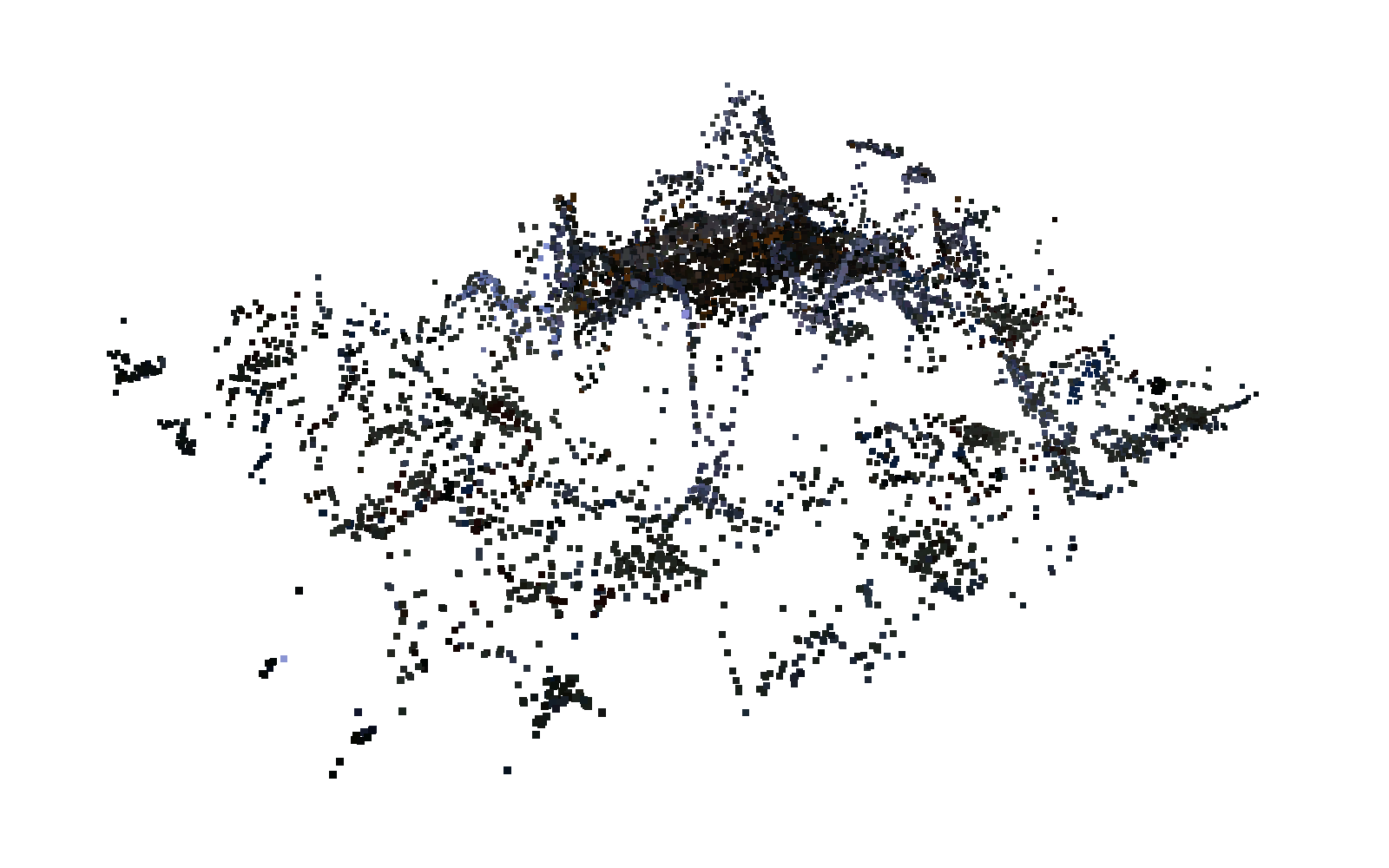}
    \end{minipage}
    (b) Moving interface with hard constraint\\
    \begin{minipage}{0.49\hsize}
        \includegraphics[width=4cm]{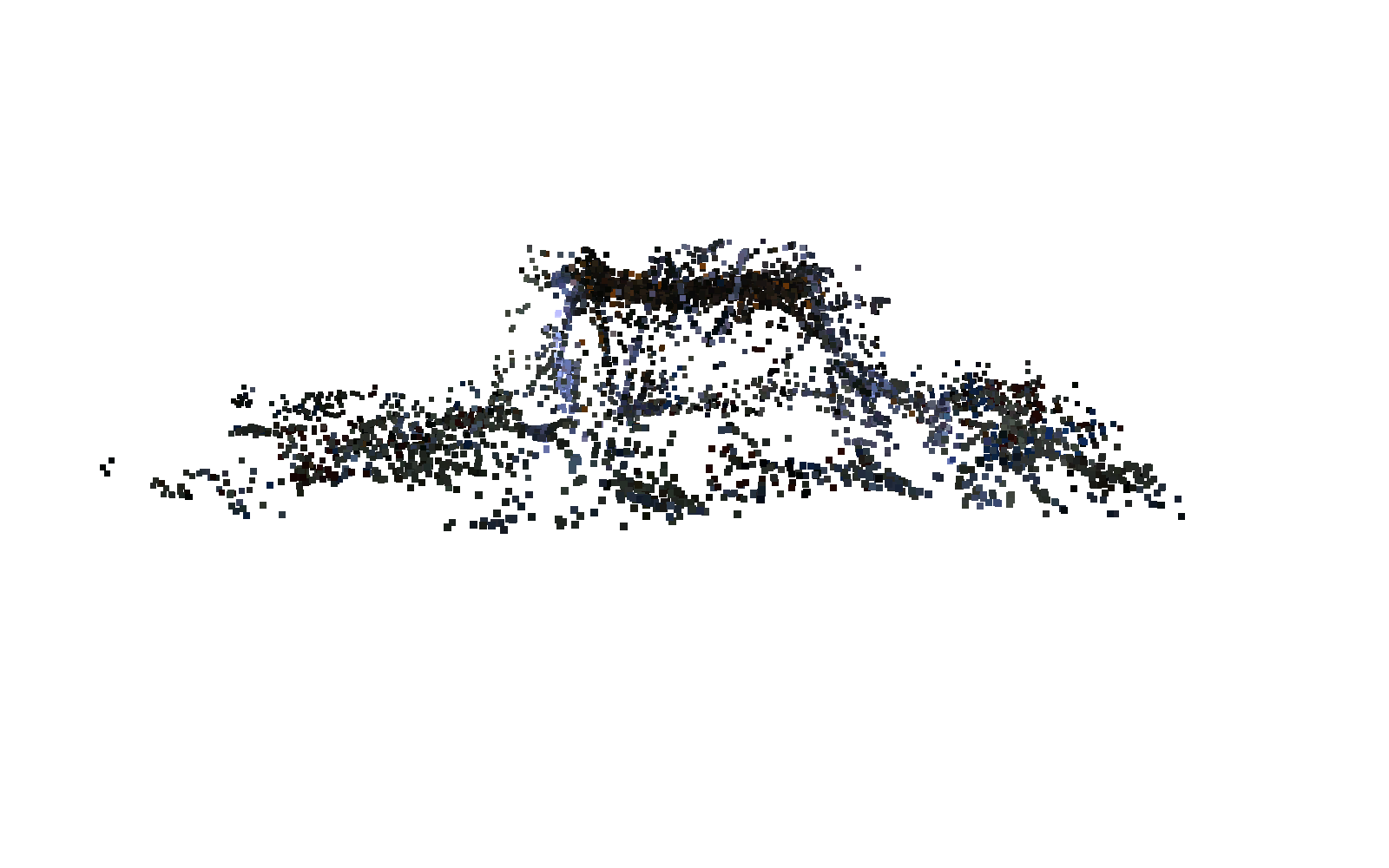}
    \end{minipage}
    \begin{minipage}{0.49\hsize}
        \includegraphics[width=4cm]{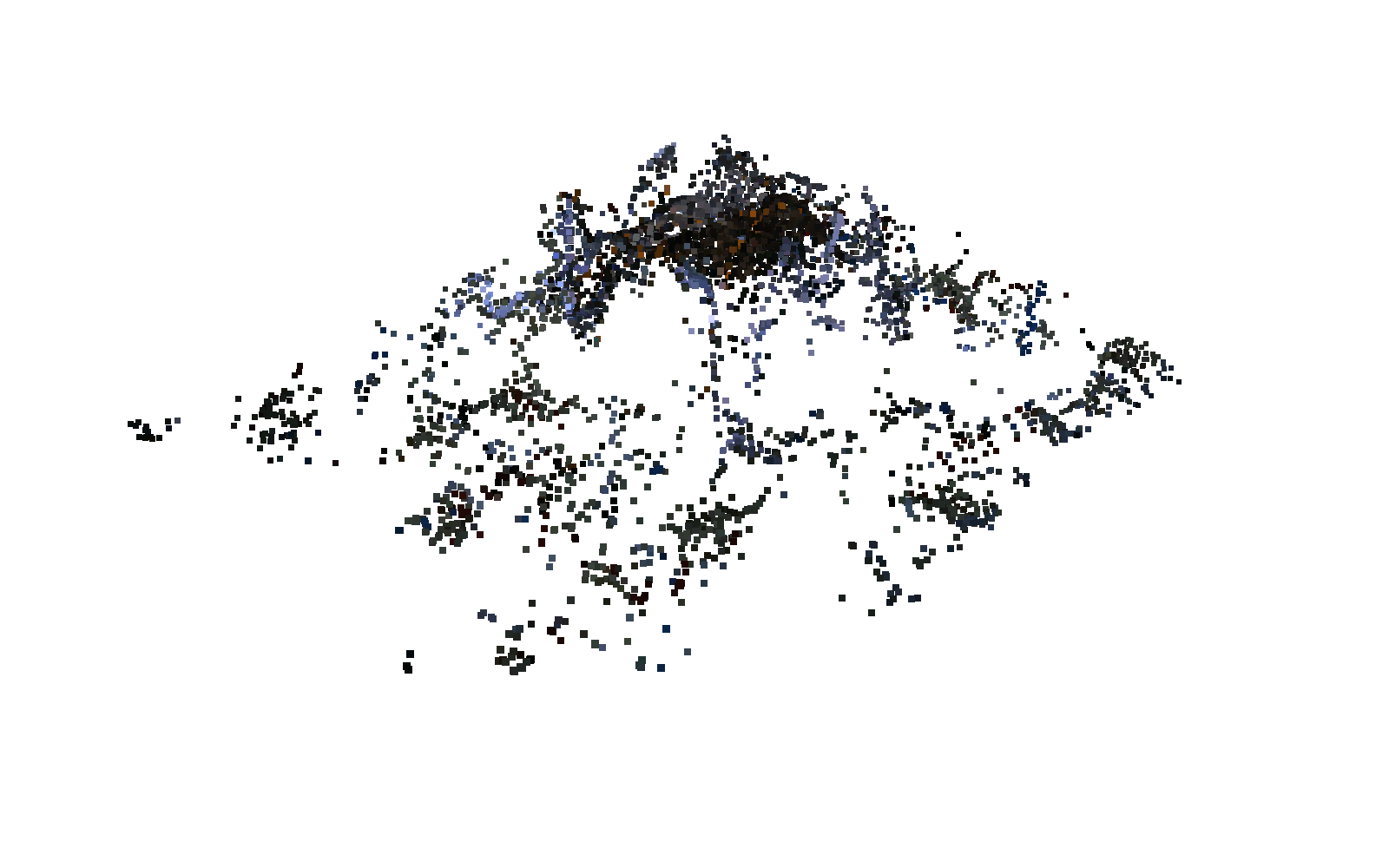}
    \end{minipage}
    (c) Moving interface with soft constraint\\
    \vspace{1mm}
    \caption{Reconstructed point clouds of the captured table and basket in static and moving interface scenarios. }
    \label{fig:table}
    \vspace{-0.5cm}
\end{figure}

Results are shown in \figref{table}.
The proposed method successfully reconstructed accurate 3D points of the table and the basket in static interface scene, and also rough shape was reconstructed in moving interface scene.
In particular, turbulence on the water surface caused many outliers in moving interface scene with hard constraint, which was successfully compensated by soft constraint, showing the effectiveness in wavy scene.

\vspace{-0.1cm}
\section{Conclusion}
\vspace{-0.1cm}
In this paper, we proposed unified underwater SfM with geometric models in three scenarios.
We provided a practical algorithm with several types of constraints used in our optimization-based reconstruction method depending on the capture setup, as well as an initialization procedure.
The experimental results showed that our method achieved higher accuracy than existing methods using real data.
We also showed that our method was able to reconstruct accurate shapes even under turbulence caused by waves.

As future work, we investigate the use of a stereo camera to cope with turbulence on wavy water surfaces, despite \cite{Qian:ECCV2018} used 9 cameras.
There might be a sufficient number of constraints to solve the unknown variables with a stereo camera if temporal information is available.
Furthermore, we consider if the stereo algorithm can be extended to SfM.

\vspace{-0.2cm}
\section*{Acknowledgment}
\vspace{-0.2cm}
K. I. and H. K. were supported by JSPS/KAKENHI 16H02849, 16KK0151, 18H04119,
18K19824, and MSRA CORE15.
\vspace{-0.2cm}

\clearpage

{\small
\bibliographystyle{ieee}
\bibliography{egbib}
}

\end{document}